\documentclass{article}

\usepackage[main,final]{neurips_2026}

\usepackage[utf8]{inputenc} 
\usepackage[T1]{fontenc}    
\usepackage{hyperref}       
\usepackage{url}            
\usepackage{booktabs}       
\usepackage{amsfonts}       
\usepackage{nicefrac}       
\usepackage{microtype}      
\usepackage{xcolor}         

\usepackage{graphicx} 
\usepackage{bm}
\usepackage{algorithm}
\usepackage{algpseudocode}
\usepackage{bbm}
\usepackage{amsthm}
\usepackage{amsmath}
\usepackage{amssymb}
\usepackage{authblk}
\usepackage{overpic}
\usepackage[capitalize]{cleveref}

\newcommand{\inp}{\mathrm{in}}
\newcommand{\out}{\mathrm{out}}

\newcommand{\net}{\mathrm{net}}

\newcommand{\T}{\mathsf{T}}
\newcommand{\C}{{\cal C}}
\newcommand{\K}{\mathcal{K}}

\newcommand{\PP}{\bm{\mathcal{P}}}
\newcommand{\nega}{-}

\title{K-prop: Deep Online Learning by Backpropagating Temporal Kernels}

\author{%
  Bingkun Liu\thanks{bingkun.liu@unibe.ch} , Paul Haider, Federico Benitez, Mihai A. Petrovici, Walter Senn\\
  Department of Physiology\\
  University of Bern\\
  Bern, Switzerland \\
}

\begin{document}
\maketitle

\begin{abstract}
    Self-recurrence represents an important component for temporal processing in neural networks, both artificial and biological, where it appears naturally in the form of leaky integration.
    However, due to the complex temporal dependencies of the gradient, an efficient and effective method of online training is still lacking.
    Here we propose K-prop, a novel online learning method for leaky neural networks.
    It dramatically reduces the computational and memory cost of real time recurrent learning (RTRL) by propagating kernel gains.
    K-prop also outperforms existing RTRL approximations and achieves performance comparable to backpropagation through time in sequence learning, classification, and reinforcement learning tasks.
    We further analyze how its sole approximation affects gradient quality and identify conditions under which K-prop remains accurate and effective.
\end{abstract}

\section{Introduction}
      
    Temporal processing is fundamental to both biological and artificial intelligence, requiring systems to transform continuous inputs into structured internal representations. Effective training for temporal dynamics relies on accurate credit assignment over time, commonly addressed by backpropagation through time (BPTT, \cite{BPTT}). However, BPTT requires storing and replaying full trajectories, limiting its applicability to online settings. In contrast, many natural and real-world systems operate continuously, where learning must occur in real time without access to stored past states, motivating the need for efficient online learning mechanisms. Prior research \citep{irie2024exploring} suggests that online learning can also improve sample efficiency in reinforcement learning settings.
    
    Artificial neural networks typically employ instantaneous neurons, whose activations depend only on the current input. In contrast, \emph{leaky (integrating) neurons} evolve according to continuous dynamics of the type $\tau \dot{u} = -u + W r$, where the membrane potential $u$ decays with time constant $\tau$ while integrating presynaptic input $r$ through weights $W$. These dynamics emulate biological membrane potentials and have therefore been widely adopted in computational neuroscience \citep{dayan2005theoretical}. Beyond biological inspiration, leaky neurons offer an intrinsic mechanism for temporal processing: each neuron implements a low-pass filter, and networks of such units naturally construct rich temporal kernels in an \emph{online} manner. Unlike Temporal Convolutional Networks (TCN, \cite{TCN}), which require access to full input sequences, leaky networks have the potential to process streaming data continuously while maintaining a distributed, multi-timescale memory, a power that has recently been demonstrated in models \citep{ELM, LiESN}. Moreover, their dynamics align naturally with analog hardware, where voltage-based circuits exhibit leaky integration \citep{stoliar2017leaky, yang2020leaky}.
 
    Despite these advantages, online training of feedforward leaky networks remains fundamentally challenging. Due to temporal integration, activity at each layer influences future downstream responses, making the gradient depend jointly on spatial pathways and temporal dynamics, analogous to recurrent networks. In principle, exact gradients can be computed by BPTT or real-time recurrent learning (RTRL, \cite{RTRL}). RTRL, while online, incurs prohibitive computational complexity of order $\mathcal{O}(n^4)$, rendering it impractical beyond very small networks. Existing RTRL approximations either reduce complexity at the expense of gradient quality or fail to scale to deep architectures. As a result, investigating online learning in large and deep leaky networks remains challenging.
    
    In this work, we introduce \emph{Kernel propagation} (K-prop), a novel learning algorithm for feedforward leaky networks that achieves efficient, online, and accurate gradient approximation. The central idea is to express gradient contributions as a decomposition over exponential temporal kernels, and to propagate only their gain coefficients backward through the network. 
    This transforms the combinatorial explosion of possible temporal pathways into a tractable, layer-wise computation analogous to standard backpropagation. 
 
    K-prop offers the following key contributions: \\
    \textbf{Efficiency and effectiveness:} K-prop significantly reduces the quartic complexity of RTRL to near-quadratic scaling, while maintaining a close approximation to the exact gradient. It achieves best-in-class  task performance, close to the exact RTRL/BPTT solutions, in deep architectures.\\
    \textbf{Continuous formulation:} Derived directly in continuous time, K-prop is robust to arbitrarily small time steps and naturally suited for analog systems and for dealing with continuous or sparse data.\\
    \textbf{Layer-wise locality:} K-prop remains layer-wise local during training, facilitating practical implementation without requiring global information pathways. 

\section{Related Work}
    A large body of work has focused on approximations to RTRL. Truncated BPTT \citep{TBPTT} reduces memory by limiting the temporal horizon, but introduces bias toward short-term dependencies. Methods that are tailored to standard recurrent networks with dense global recurrence do not align well with leaky neurons, whose dominant temporal effect arises from strong \emph{self-recurrence} (diagonal dynamics). For example, low-rank approximations such as unbiased online recurrent optimization (UORO, \cite{UORO}) fail to capture the inherently high-rank diagonal structure, while Kronecker-based methods \citep{KF-RTRL, OK-RTRL} do not exploit the naturally separable temporal kernels. Moreover, these approaches are formulated in discrete time and are not directly suited to continuous-time or analog dynamics.
    
    Alternatively, methods such as random feedback local online learning (RFLO, \cite{RFLO}) and e-prop \citep{eprop} are specifically designed for leaky neurons and emphasize biological plausibility. These methods rely on local eligibility traces but discard inter-neuron dependencies, which prevents effective credit assignment in deep layers. 
    \cite{zucchet2023online} derived online learning for complex-valued leaky-like memory, and is also limited to a single memory layer.
    Online spatio-temporal learning (OSTL, \cite{OSTL}), partially addresses this by propagating instantaneous errors to deep layers whose temporal dependencies are still missing.
    Following an energy-based approach, generalized latent equilibrium (GLE, \cite{GLE}) assumes a sufficiently smooth error signal $\epsilon$, so that the prospective error $\breve{\epsilon} = \epsilon +\tau\dot{\epsilon}$ can be used to estimate temporal gradients. 
    Beyond previous online methods, K-prop preserves the temporal structure of the deep layer gradient by deploying a novel decompositions of both temporal kernels and spatial paths.

\section{Theory}

    \subsection{K-prop: Propagating kernel gains}
    
\begin{figure}[tbp]
            \centering
            \begin{overpic}[width=0.35\linewidth]{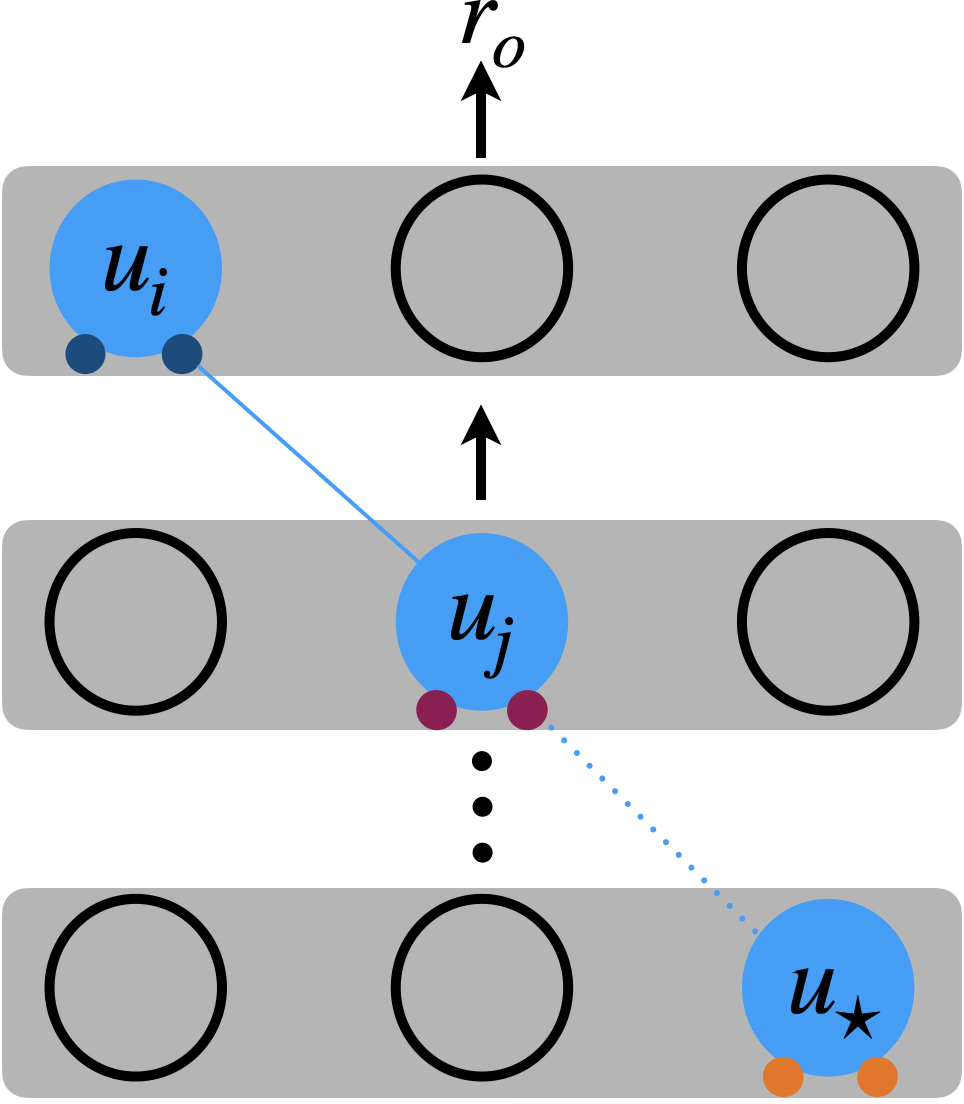}
                \put(-2, 93) {\textbf{a}} 
            \end{overpic}
            \begin{overpic}[width=0.6\linewidth, trim=0.2cm 0.2cm 0.2cm 0cm,clip]{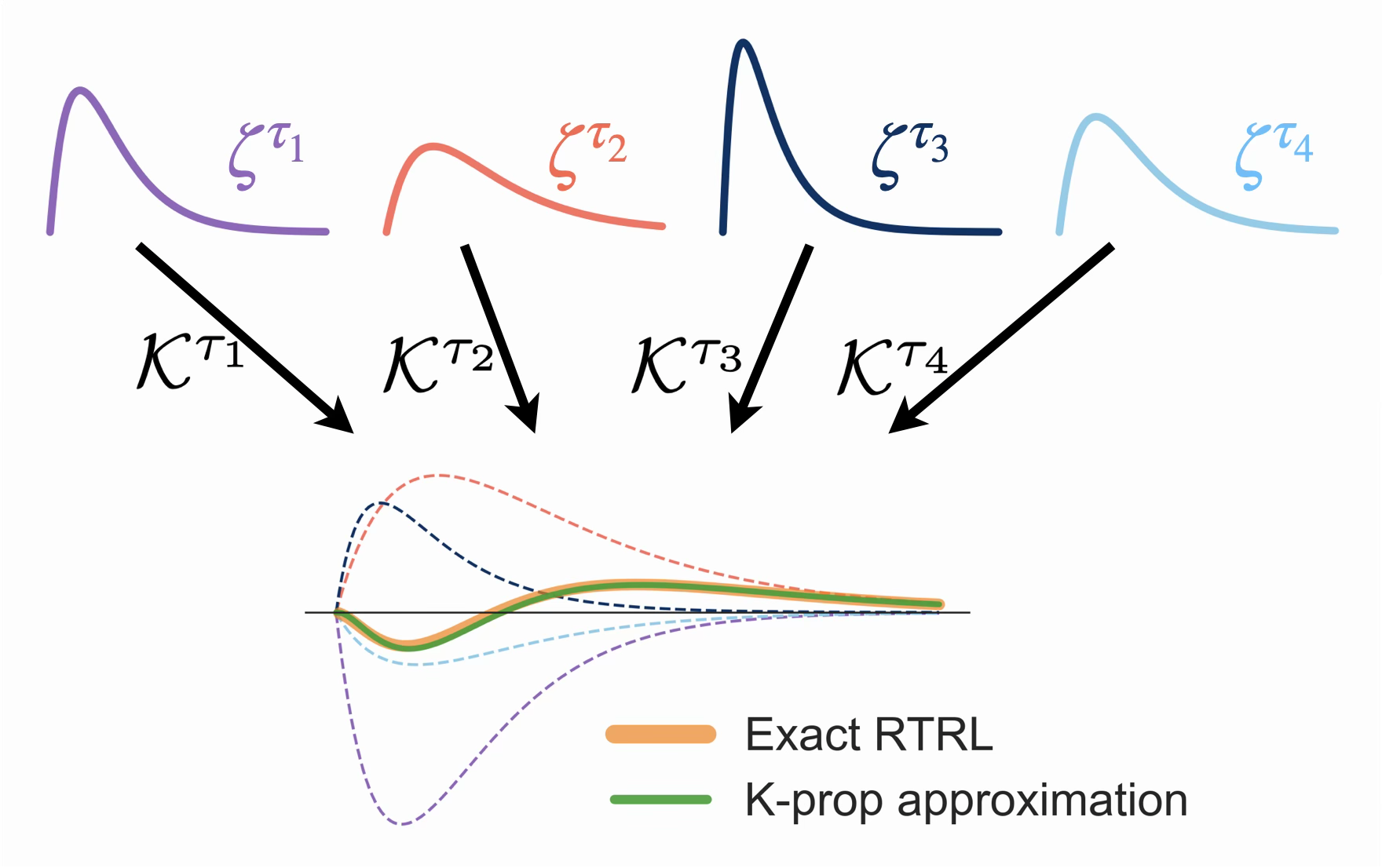}
                \put(0, 62) {\textbf{b}}
            \end{overpic}
            \caption{Kernel propagation.
            (a) Calculating gradients involves hierarchical filters along paths of neuron connections.
            (b) Illustration of kernel decomposition.
            The complicated temporal structure in the true gradient can be approximated by a sum over $\zeta$ weighted by $\tau$-dependent coefficients $\K$. Every $\zeta$ (solid) is weighted by corresponding $\K$ as a component (dashed in the same color). The sum of all components aligns well with the exact term computed by RTRL. 
            }
            \label{fig:path}
        \end{figure}

        We consider a feedforward leaky neural network with nonlinear activation $r=\rho(u)$.
        From the output loss function $\mathcal{C}(t)$, gradients propagate backward to each hidden neuron through the downstream network along all possible directed paths (as shown in \cref{fig:path}a). 
        The gradient of each path constains a \emph{temporal term} accounting for its temporal contribution. 
        For a postsynaptic neuron $i$ in the last leaky layer $L$, this temporal term is represented by a vector of synaptic eligibility traces $\bm z_i =\rho'_i \overline {\bm r}^{\tau_i}_{\![L-1]}$,
        where $ \overline{\bm x}^{\tau} = \kappa^{\tau} * \bm x$ denotes a low-pass filtering of the vector-valued temporal variable $\bm x_t$ with the exponential kernel $\kappa^{\tau} = \Theta(t) \frac1{\tau}\exp(-\frac{t}{\tau})$, where $\Theta$ is the Heaviside function.
        Moving one layer upstream to neuron $j$ introduces an additional filtering $\rho'_i(\kappa^{\tau_i} *\bm z_j)$, and an arbitrary neuron $\star$
        in deeper layers generates more iterative hierarchical filters  $\rho'_i(\kappa^{\tau_i} *\rho'_j(\kappa^{\tau_j}*...*\bm z_\star))$.
        Therefore, each path induces a \emph{hierarchical temporal kernel} characterized by the leaky dynamics of all neurons along the path, and the exact gradient requires summing over all combinations of such path-dependent kernels. 
        
        K-prop simplifies this involved computation in three steps.
        
        \textbf{Step 1: Separating nonlinearity from filtering}
        
            K-prop approximates $\rho'$ by a surrogate $\hat{\rho'}$ ($=  \overline{ \rho'_\star}^{\tau_\star}$ as default), such that it can be moved outside of the filter: $\overline{\rho'x}^{\tau} \approx \hat{\rho'} \overline{x}^{\tau}$.
            We can therefore approximate a hierarchical filter by a nested filter: $\rho'_i(\kappa^{\tau_i}*\rho'_j(\kappa^{\tau_j}*\rho'_k(\kappa^{\tau_k}*...)))  \approx (\rho'_i \hat{\rho'_j}\hat{\rho'_k}...)(\kappa^{\tau_i} * \kappa^{\tau_j}*\kappa^{\tau_k}*...)$. The assumption that $\rho'$ varies slowly relative to the filtering time scale $\tau$ is the only approximation required by K-prop. 
        
        \textbf{Step 2: Collapsing nested filters into basis kernels}
        
            \fbox{
                \noindent
                \begin{minipage}{0.9\linewidth}
                    \textbf{Lemma 1}: Nested exponential filters can be rewritten as a linear combination of single filters. $\Upsilon$ is a set of N mutually distinct $\tau$. A proof is given in \cref{sec:lemmaproof}.
                    \begin{align}
                       \kappa^{\Upsilon}(x) &=  \kappa^{\tau_0} * \kappa^{\tau_1}*\ldots\kappa^{\tau_N} * x
                        \\
                        &= \sum_{\tau_i\in {\Upsilon}} 
                        c_{i} (\kappa^{\tau_i} *x) \mbox{ with  }  c_{i}= \prod_{
                    \tau_j\in {\Upsilon},j \neq i}
                   \frac{\tau_i}{\tau_i -\tau_j} \, .
                        \label{eq:mainHierLP} 
                    \end{align}
                \end{minipage}
            }
            
            In this work, we treat identical $\tau$ across layers as slightly perturbed, which yields an equivalent temporal horizon while avoiding singularities in \cref{eq:mainHierLP}.
            Consequently, each $\tau$ appears once in a path. Applying the lemma, each \emph{path-specific nested kernel} is converted into a \emph{linear combination of basis kernels}.
            Together with Step 1, the path gradient can then be approximated as $
            \frac{\partial \mathcal{C}(t)}{\partial W_\star}\Big|_p
            \;\approx\;
            w_p \sum_{i\in p^{\setminus \star}} c_i (\kappa^{\tau_i} * \bm z_\star)$, which sums over all neurons $i$ in the path $p$ except the initial $\star$.
            The term $w_p$ is the \textbf{path weight}, a product of all synaptic weights, nonlinear surrogates $\hat{\rho'}$, and the output error along path $p$ (see \cref{eq:weightpath} in Supplement).
           
        \textbf{Step 3: Propagating kernel gains through layers}
        
            Despite the reduction above, explicitly summing over paths remains intractable given: (1) exponential growth in the number of paths;
            (2) dependence on detailed network topology; (3) dependence on global downstream variables (weights, activities, errors).
            The key observation is that the number of distinct time constants is much smaller than the number of paths.
            Let $\mathcal{P}$ denote the set of all downstream paths and $\Upsilon_\mathrm{down}$ represent the set of all distinct time constants in downstream neurons.
            We can rearrange the summation over paths and group coefficients by their shared time constant, $\sum_{p \in \mathcal{P}_\star}
            w_p \sum_{ i\in p} c_i \, (\kappa^{\tau_i} * \bm z_\star) = \sum_{\tau \in \Upsilon_\mathrm{down}^\star}
            \mathcal{K}^\tau_\star \, (\kappa^{\tau} * \bm z_\star)$, where $\mathcal{K}^\tau_\star$ is a \textbf{kernel gain} that aggregates \emph{all path contributions} associated with time constant $\tau$. 
            This transformation converts the sum over
            \emph{all combinations of paths and their neurons} to a sum over \emph{a small set of kernels}. 
            
            \begin{figure}[tbp]
                \centering
                \includegraphics[width=0.98\linewidth, trim=0cm 0cm 0cm 0cm,clip]{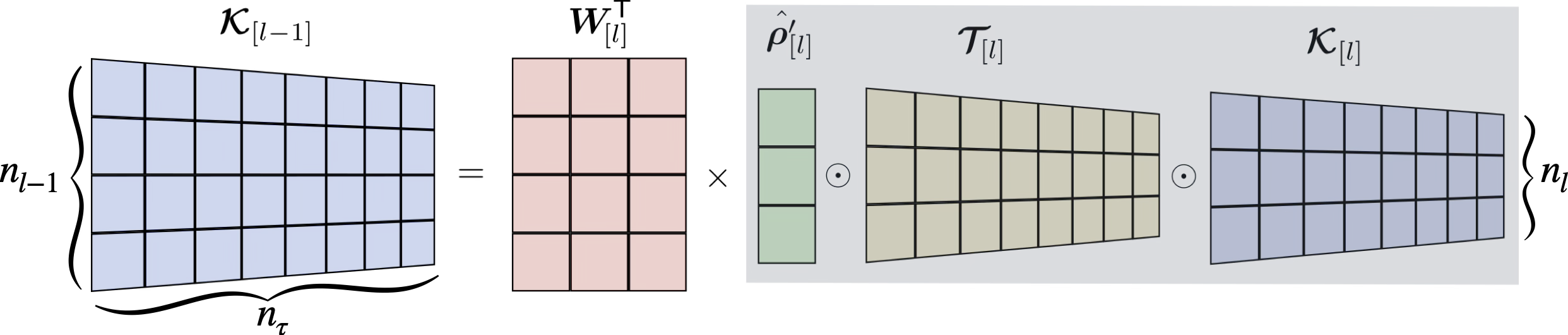}
                \caption{Kernel propagation operation. Example propagation from layer $l$ with 3 neurons to layer $l-1$ with 4 neurons. The gray box works as a bracket.}
                \label{fig:kernelBP}
            \end{figure}
            Another central insight of K-prop is that $\mathcal{K}$, which implicitly depends on all downstream paths, satisfies a \emph{layer-wise recursive relation}, i.e.,
            $\K^\tau_{[l-1]}$ can be derived from $\K^\tau_{[l]}$, in a way analogous to backpropagation. For a compact notation,  we define $\Upsilon_\net$ as the set of all distinct $\tau$ in the network, and define $\tau$-related transformation factors at neuron $j$:
            \begin{equation}
                \mathcal{T}^{\tau}_{j} = 
                \begin{cases}
                    1& \tau_j=\tau\\
                    0&  \text{$\tau$ in the same layer as neuron $j$ \text{ and } $ \tau \neq \tau_j$} \\
                    \frac{\tau}{\tau -\tau_j} & \text{otherwise} \, .
                \end{cases}
                \label{eq:T}
            \end{equation}
            This definition for $\mathcal{T}$ is based on fully connected feedforward networks, and is further elaborated and extended in \cref{sec:T}.
            With $n_\tau = |\Upsilon_\net|$, let $\boldsymbol{\mathcal K}_{[l]} \in \mathbb{R}^{n_l \times n_\tau}$ denote the kernel gain at layer $l$, $\bm{\mathcal{T}}_{\!\![l]} \in \mathbb{R}^{n_l \times n_\tau}$ the corresponding transformation matrix, and $\hat{\bm \rho'}_{\![l]} \in \mathbb{R}^{n_l}$. The external error captured by the loss enters from the output layer as the gain $\boldsymbol{\mathcal K}^\tau_o = \frac{\partial \mathcal{C}}{\partial \bm r_o}$, and $\bm \K$ propagates backward according to (\cref{fig:kernelBP})
            \begin{align}
                \boldsymbol{\mathcal{K}}^{\tau}_{[l-1]} = \boldsymbol{W}_{\![ l]}^\intercal (\hat{\boldsymbol{\rho}'}_{\![l]}
               \odot \boldsymbol{\mathcal T}^\tau_{\!\![l]} \odot \boldsymbol{\mathcal{K}}^\tau_{[l]}) \, .
            \label{eq:K-propagation}
            \end{align}
            We consider the eligibility matrix $\boldsymbol{\mathcal{Z}}_{[l]}   \in \mathbb{R}^{n_l \times n_{l-1}}$ for layer $l$, with the eligibility vector $\bm z_{[l], i}^{\intercal}$ appearing in the $i$'th row.
            With this we define \textbf{deep eligibility traces} $\boldsymbol{\zeta}^{\tau_d}_{[l]} = \overline{\boldsymbol{\mathcal{Z}}_{[l]}}^{\tau_d} \in \mathbb{R}^{n_l \times n_{l-1}}$  for all downstream time constants $\tau_d \in \Upsilon_\mathrm{down}^l$.
            Each synapse thus tracks a number of eligibility traces equal to the number of different neuronal time constants that appear in the downstream network. This can be done locally and online through simple leaky integrator dynamics in a matrix form:
            \begin{equation}
               \tau_d \, \dot{\bm \zeta}^{\tau_d}_{[l]} = - \bm \zeta^{\tau_d}  _{[l]} + \boldsymbol{\mathcal{Z}}_{[l]}\,.
               \label{eq:xi}
            \end{equation}
            Finally, the total gradient sums up the contribution from all kernel components
            \begin{align}
                \dot{\boldsymbol{W}}_{\![l]} \propto \sum_{\tau \in \Upsilon_\mathrm{down}^l} \text{diag}(\boldsymbol{\mathcal{K}}_{[l]}^{\tau}) \, \boldsymbol{\zeta}_{[l]}^{\tau} \, .
            \end{align}
            K-prop replaces explicit path enumeration by propagating kernel coefficients. Each neuron tracks local temporal features $\bm \zeta$, while $\boldsymbol{\mathcal{K}}$ carries error information across layers (\cref{fig:path}b). This yields a fully online and continuous time learning rule that avoids combinatorial complexity while preserving the structure of the path-based gradient (see pseudocode in \cref{alg:kp_online}).           
            In order to enable this mechanism, global information is used at initialization to construct $\boldsymbol{\mathcal T}$ and $\Upsilon_{\mathrm{down}}$. During training, all computations remain layer-wise local. Consequently,  K-prop can be straightforward to implement by integrating feed-forward network techniques, treating $\bm \K$ as the gradient in standard backpropagation. Normalization or dropout simply scales or drops out $\bm \K$, and convolutional or residual architectures only change the matrix operation during propagation.

    \subsection{Quality of the gradient approximation}

        The accuracy of K-prop is governed by its sole approximation $\frac{\rho'(t' <t)}{\hat\rho_{'}(t)} \approx 1$, which assumes that $\rho'$ varies slowly over the relevant timescales determined by the inputs.
        Deviations from this assumption introduce a mismatch between the true gradient and its K-prop estimate. Understanding how this mismatch depends on network and model properties is key to improving K-prop performance.
        
        We quantify gradient quality using the cosine dissimilarity between K-prop and BPTT gradients:
        \[
        1 - \frac{\Delta W_{\mathrm{KP}} \cdot \Delta W_{\mathrm{BP}}}{\|\Delta W_{\mathrm{KP}}\| \, \|\Delta W_{\mathrm{BP}}\|}.
        \]
        Across configurations, the median dissimilarity remains low ($<10^{-2}$), indicating that most neurons receive well-aligned gradients.
        
        \begin{figure}[tbp]
            \centering
            \begin{overpic}[width=0.3251\linewidth]{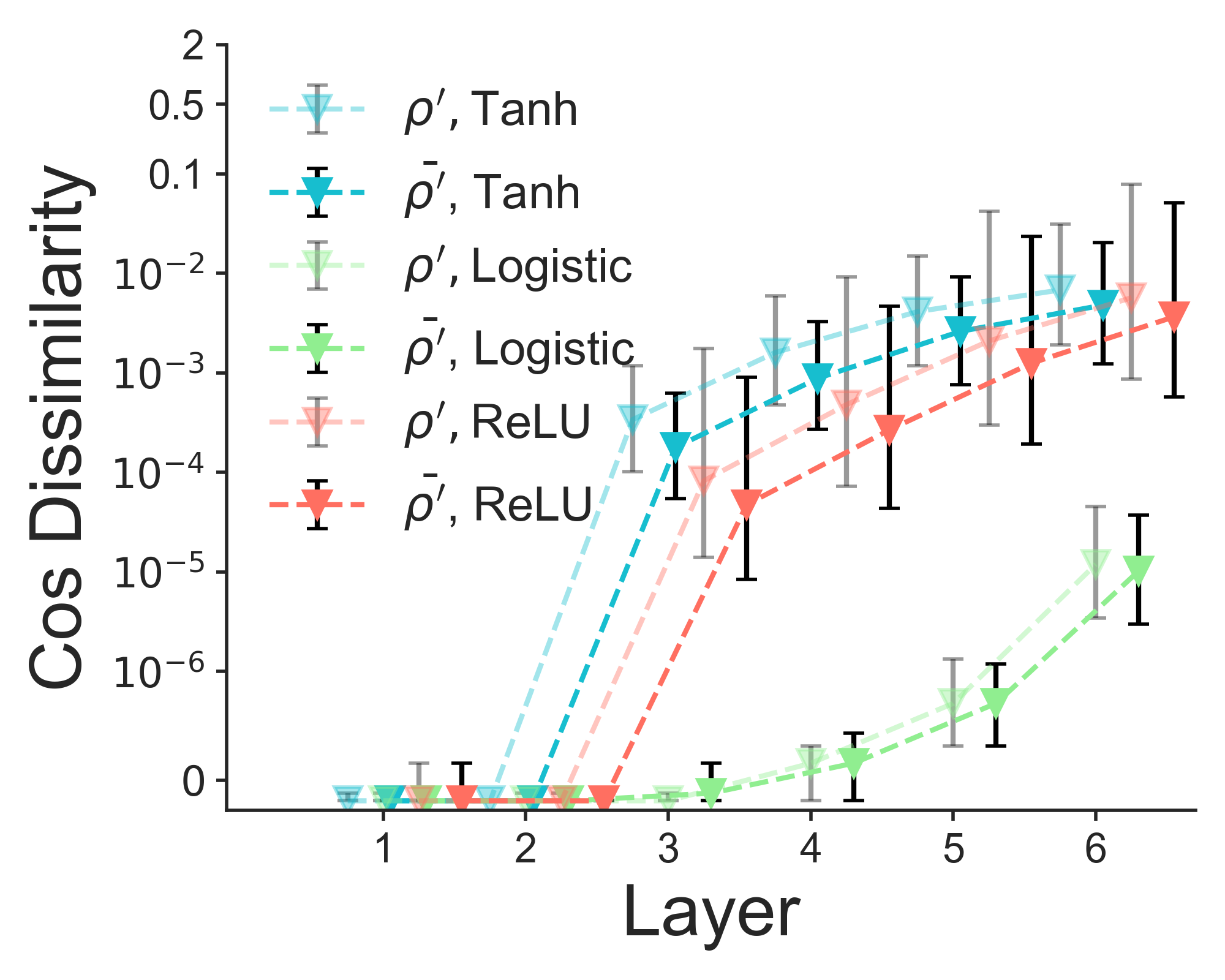}
                \put(3, 70) {\textbf{a}} 
            \end{overpic}
            \begin{overpic}[width=0.3251\linewidth]{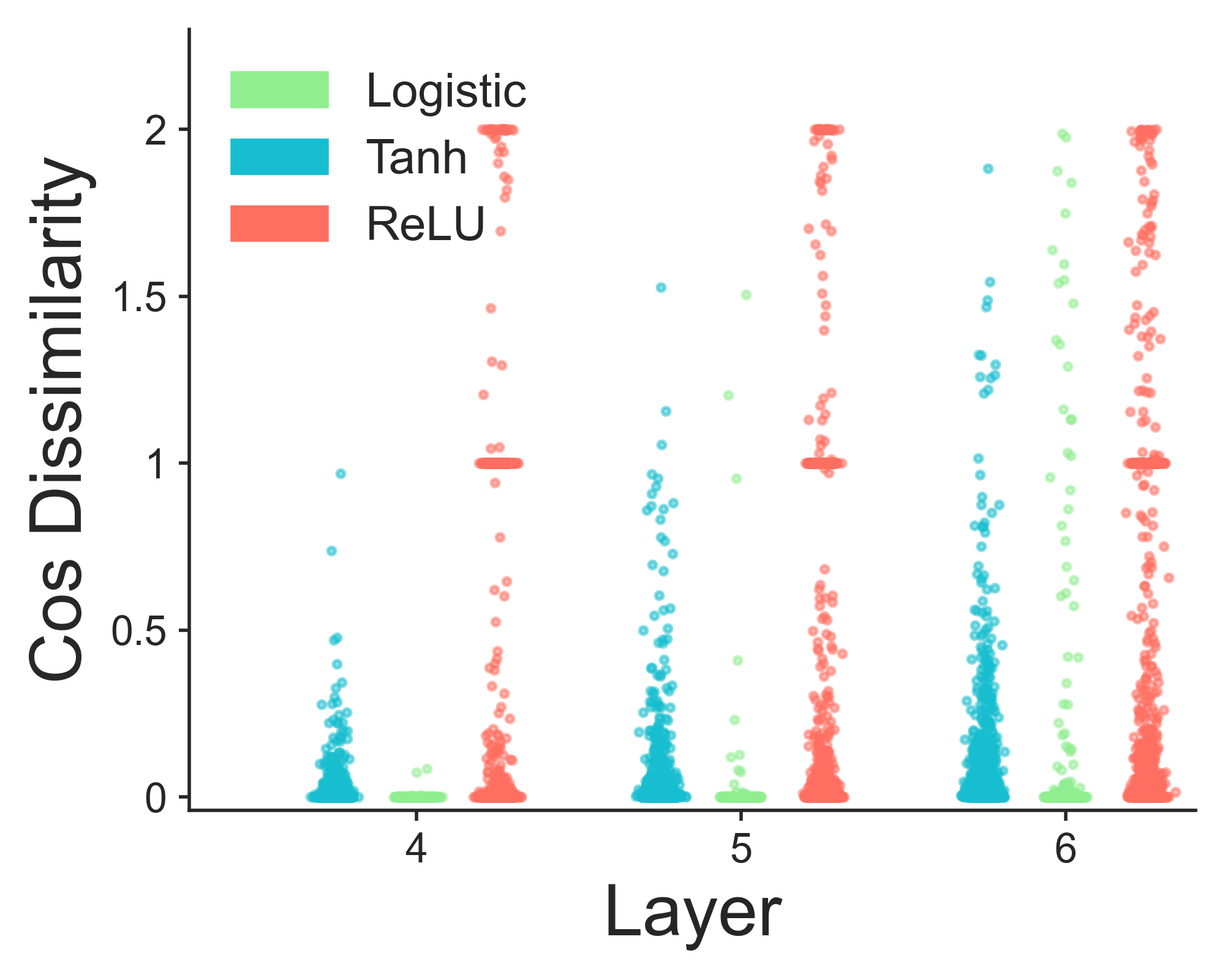}
                \put(2, 70) {\textbf{b}}
            \end{overpic}
            \begin{overpic}[width=0.3251\linewidth]{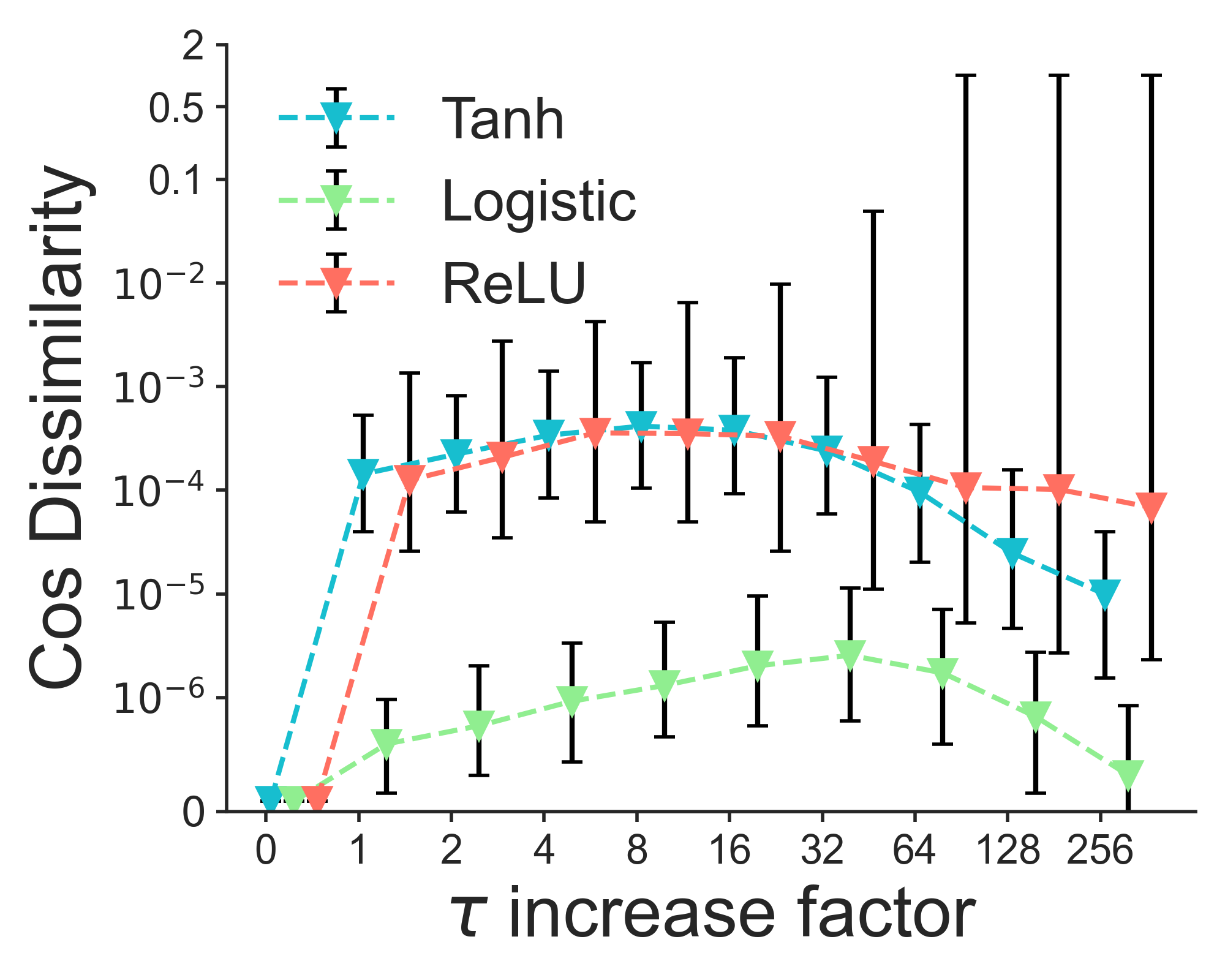}
                \put(3, 70) {\textbf{c}}
            \end{overpic}
            \caption{Gradient cosine dissimilarity between K-prop and exact (BPTT/RTRL) gradients in randomly initialized networks, measured with respect to each neuron given a sample of the MNIST1D dataset (see \cref{Sec:MNIST}).
            Dissimilarity ranges from $0$ to $2$ with $0$ indicating complete alignment, $1$ orthogonality, and $2$ complete anti-alignment. Error bars show median, 1st, and 3rd quartiles.
            (a) Gradient mismatch increases towards earlier layers, as an increasing number of $\rho'$ need to be approximated as constant.
            The mismatch depends on the type of nonlinearity present in the network, and $\bar{\rho'}$ works better than $\rho'$ as surrogate $\hat{\rho'}$.
            (b) Scatter plots show the distribution of outliers. ReLU leads to more outlier neurons than sigmoid nonlinearities.
            (c) Gradient mismatch increases, then decreases for larger $\tau$.
            All plots show sigmoid functions performing better than ReLU.
            }            
            \label{fig:GradientQuality}
        \end{figure}        
        \textbf{Depth.}
        As shown in \cref{fig:GradientQuality}a, gradient dissimilarity increases with distance from the output layer. The last layers are computed exactly, while upstream layers accumulate approximation errors
        through repeated propagation.
        This is consistent with the layer-wise structure of K-prop, where small local mismatches compound along depth.
        
        \textbf{Time constants.}
        The approximation error is directly linked to the timescales of the input (and thus the network) variation.
        When all $\tau=0$, K-prop reduces to standard backpropagation
        and is exact. As $\tau$ increases, dissimilarity grows due to longer temporal integration windows.
        In the limit of very large $\tau$, network dynamics become slower, reducing temporal variation in $\rho'$ and improving the K-prop approximation. Note that this limit only serves as a cross-check for our intuition: while gains $\rho'$ become more constant over time, networks also lose functionality as they become incapable of meaningfully processing incoming signals.
        
        \textbf{Nonlinearity.}
        The choice of activation function determines the smoothness of $\rho'$. Functions with slowly varying derivatives (e.g., logistic sigmoid) yield lower dissimilarity. Although $\tanh$ is an offset and scaled logistic function, its derivative has larger magnitude and varies more rapidly, leading to increased mismatch. ReLU presents a distinct regime: $\rho'$ is piecewise constant but discontinuous at zero. As a result, gradients are either highly accurate or highly inaccurate, leading to low median error but significantly higher variance, as reflected by the outlier distribution.
        
        \textbf{Surrogate derivative.}
        The design of $\hat{\rho'}$ directly controls the approximation. 
        For example, a low-pass-filtered version $\bar{\rho_{'}}^{\tau_\rho}(t)$ improves gradient alignment compared to using the instantaneous $\rho'(t)$, and the $\tau_\rho$ used to filter $\rho'$ shifts the gradient alignment tradeoff across network depth (\cref{sec:surrogate}).
        Exploring surrogate replacements is a promising direction for further improving K-prop.
        
        Overall, these results show that gradient quality in K-prop is determined by the interaction between temporal dynamics and nonlinearity, and can be systematically improved through appropriate choices of network timescales, activation functions, and surrogate design.
  
\section{Experiments}
    \subsection{Continuous signal processing with memory}

        \begin{figure}[t]
            \centering
            \begin{overpic}[width=0.4\linewidth]{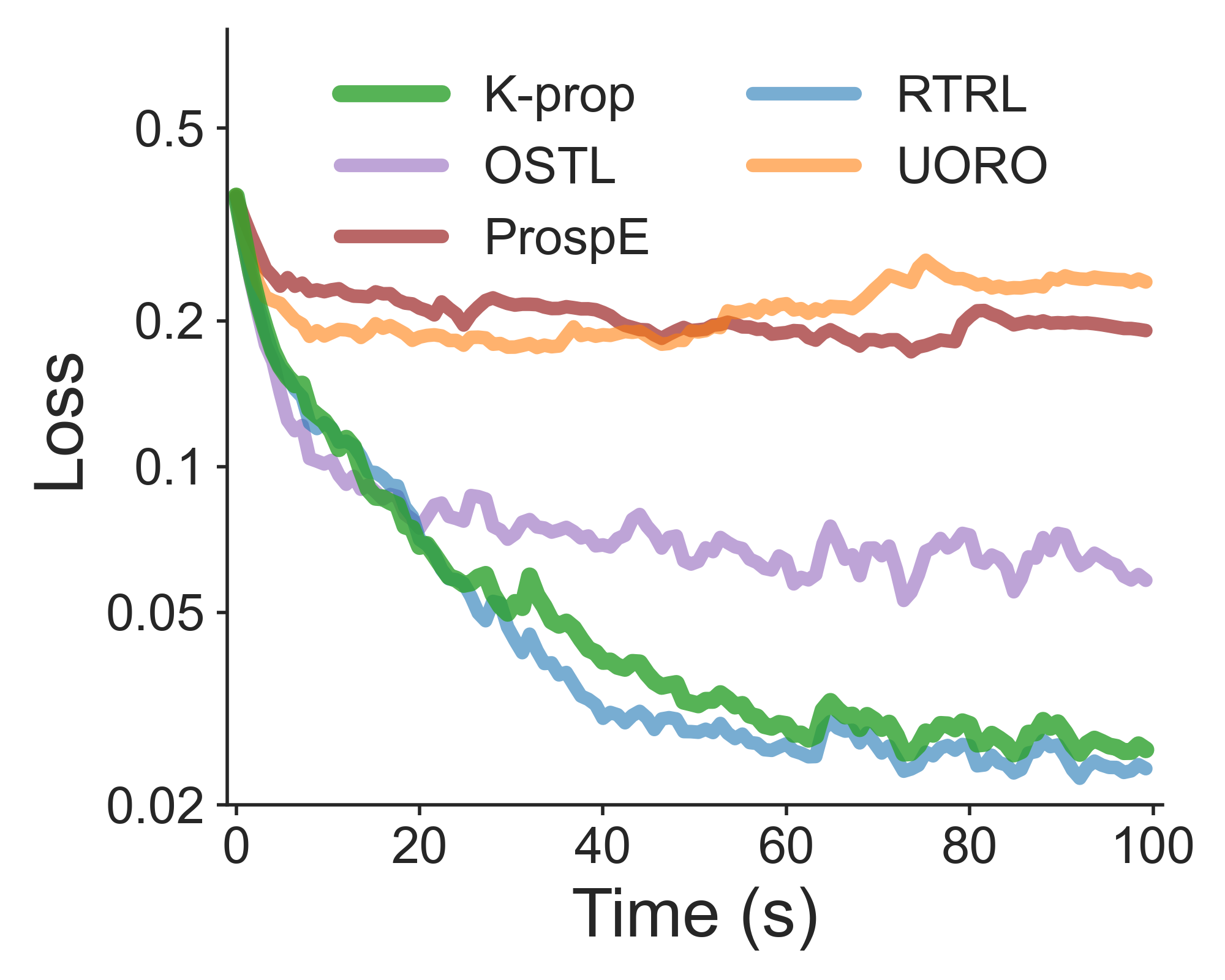}
                \put(4, 65) {\textbf{a}} 
            \end{overpic}
            \begin{overpic}[width=0.45\linewidth]{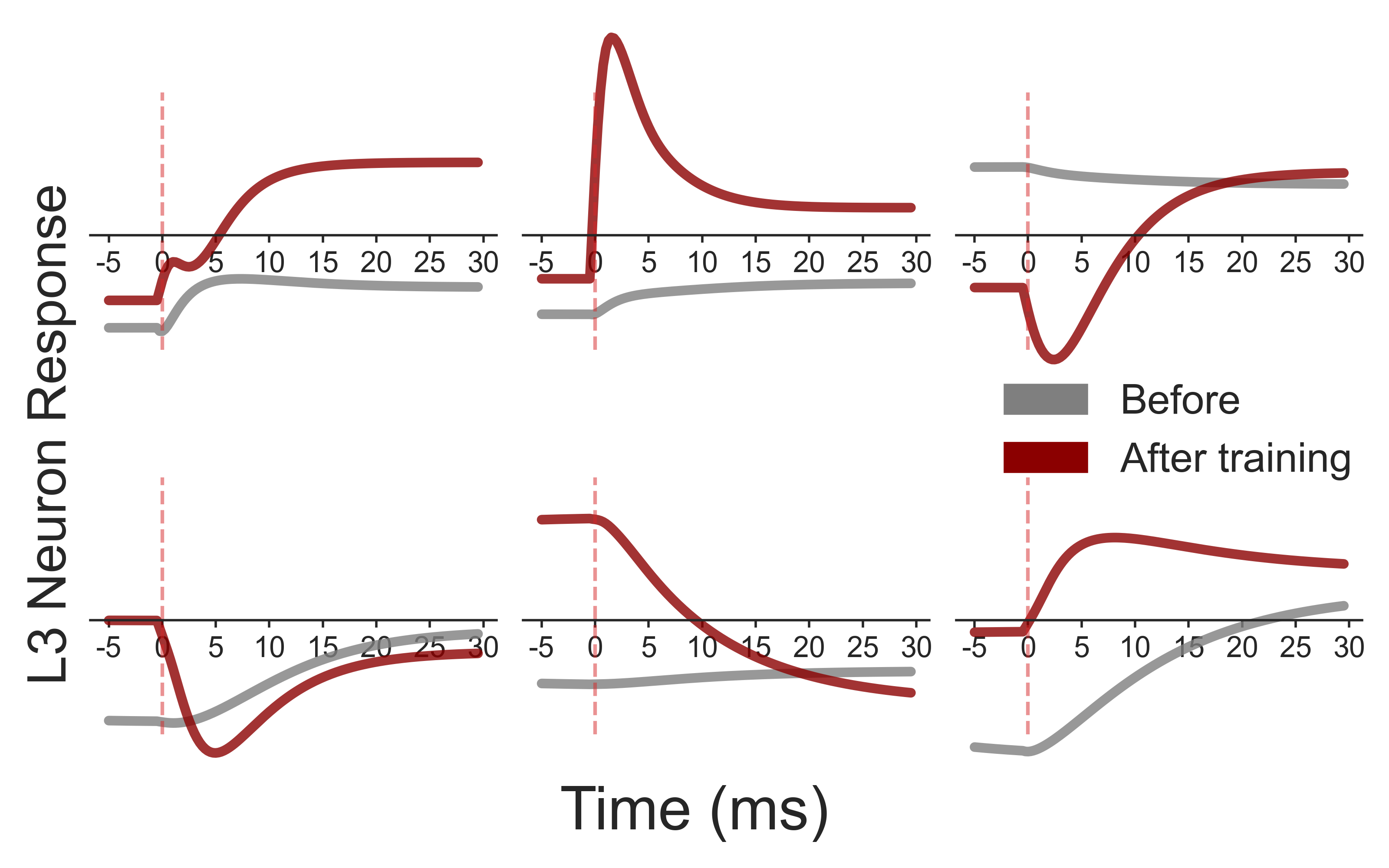}
                \put(2, 52) {\textbf{b}}
            \end{overpic}
            \begin{overpic}[width=0.8\linewidth]{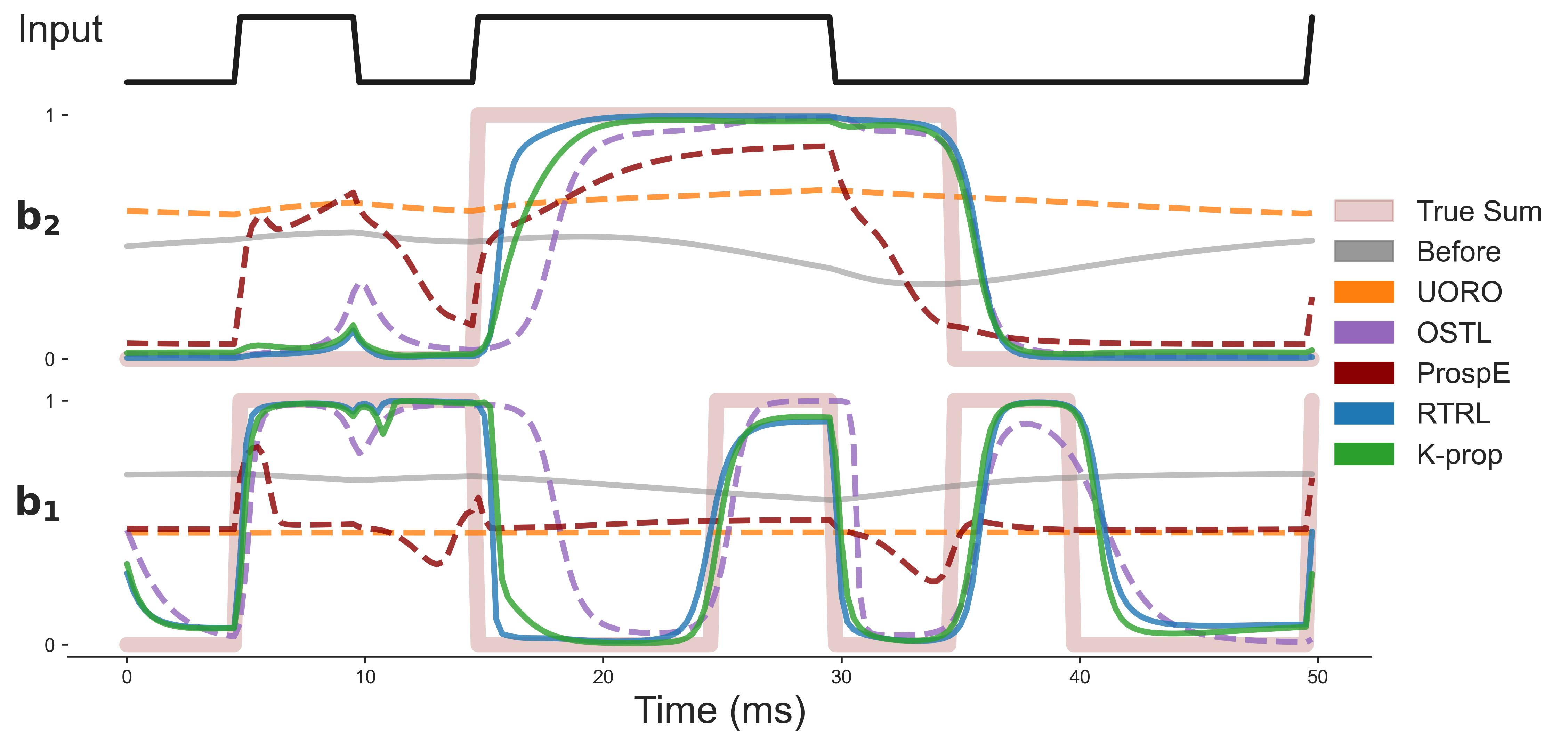}
                \put(-2, 45) {\textbf{c}}
            \end{overpic}
            \caption{N-sum task evaluated with various online learning methods.
            ProspE only considers prospective errors $\breve{\epsilon}$ in GLE \citep{GLE}, without further augmentations.
            (a) Training loss over time. K-prop achieves substantially lower error than other approximation methods and closely matches RTRL.
            (b) Temporal response of hidden neurons to a step input before and after learning. K-prop develops more structured and contrastive temporal kernels. (c) Example input and corresponding outputs.
            Before learning, all models produce non-informative outputs.
            After learning, K-prop and RTRL closely align with the target sequence, while other methods deviate in both timing and amplitude. 
            }
            \label{fig:N-Sum}
        \end{figure}

        We first evaluate K-prop on a clearly defined non-episodic task, the N-sum problem with $N=3$. The model receives a continuous binary stream (e.g., [0, 1, 1, 1, 0, ...]) without truncation and must output the sum of the most recent three bits in binary form using two output units ($b_2 b_1 \in \{00,01,10,11\}$). Each bit is presented for 5 ms, requiring the model to maintain a memory of at least $5N=15$ ms. Note that the time units are, in principle, arbitrary, and what matters is the ratio between time constants and bit duration.
        
        Despite its simple formulation, the task presents two key challenges: (1) the model is not informed of the temporal horizon $N$ or the bit duration, and must infer the relevant memory timescale purely from delayed error signals; (2) the least significant output bit $b_1$ effectively implements a parity function over recent inputs, which is a class of problems known to be difficult for gradient-based learning \citep{parity1, parity2}.

        The loss is defined as the mean squared error and is delivered 2 ms after the onset of each input bit (reaction latency).
        Gradients are computed online at every time step, accumulated into $\Delta W$, and applied periodically, after which the accumulator is reset (see weight update period in \cref{sec:weightupdate} and Supplement ~\cref{table:networkconfig}). This update scheme is shared across all methods.

        To emphasize learning under constrained capacity, we use a small architecture (6-6-6-2) with logistic output units. In this setting, exact RTRL remains tractable, allowing for a controlled comparison where all methods share identical initialization and learning rates. K-prop achieves performance nearly indistinguishable from RTRL, while significantly outperforming other online approximation methods. The advantage becomes more pronounced on the more difficult parity component $b_1$, where K-prop better matches the target in both timing and amplitude (\cref{fig:N-Sum}). In N-sum, the correct output depends on integrating information over a precise temporal window ($\sim$15 ms) while simultaneously performing a nonlinear transformation (parity-like mapping). This requires assigning credit not only locally in time, but also across combinations of temporal filters. 

        Beyond output accuracy, we analyze the temporal responses of hidden neurons (layer L3) in \cref{fig:N-Sum}. During training, neurons develop diverse temporal kernels that extract task-relevant features. Compared to initialization, the learned responses to step inputs become both more expressive and temporally structured. For instance, neurons exhibiting sign changes around 10 ms, effectively compute differences between recent and earlier inputs, resembling temporal filtering operations. Such structured temporal responses have been widely observed in biological systems, including retina \citep{trenholm2020annotated}, visual cortex \citep{Zhaoping2014}, and auditory system \citep{lesica2008efficient, zhao2011understanding}, suggesting convergent strategies for temporal feature extraction.

        \begin{table}[t]
            \centering
            \caption{Test accuracy (\%) of BPTT, OSTL, and K-Prop across the evaluated datasets. Results are reported as mean $\pm$ standard deviation.}
            \label{tab:accuracy_comparison}
            \begin{tabular}{lccc}
                \toprule
                Method & MNIST1d-Lk4 & MNIST1d-Lk5 & GSC \\
                \midrule
                BPTT \cite{BPTT}  & $96.60 \pm 0.20$& $97.69 \pm 0.21$& $92.47 \pm 0.21$\\
                OSTL  \cite{OSTL} & $96.08 \pm 0.30$& $97.03 \pm 0.16$& $90.61 \pm 0.20$\\
                \textbf{K-prop (Ours)} & $\mathbf{96.53\pm0.54}$& $\mathbf{97.56\pm0.19}$& $\mathbf{91.52\pm0.10}$\\
                \bottomrule
            \end{tabular}
        \end{table}

        \begin{figure}[tbp]
            \centering
            \begin{overpic}[width=0.32\linewidth]{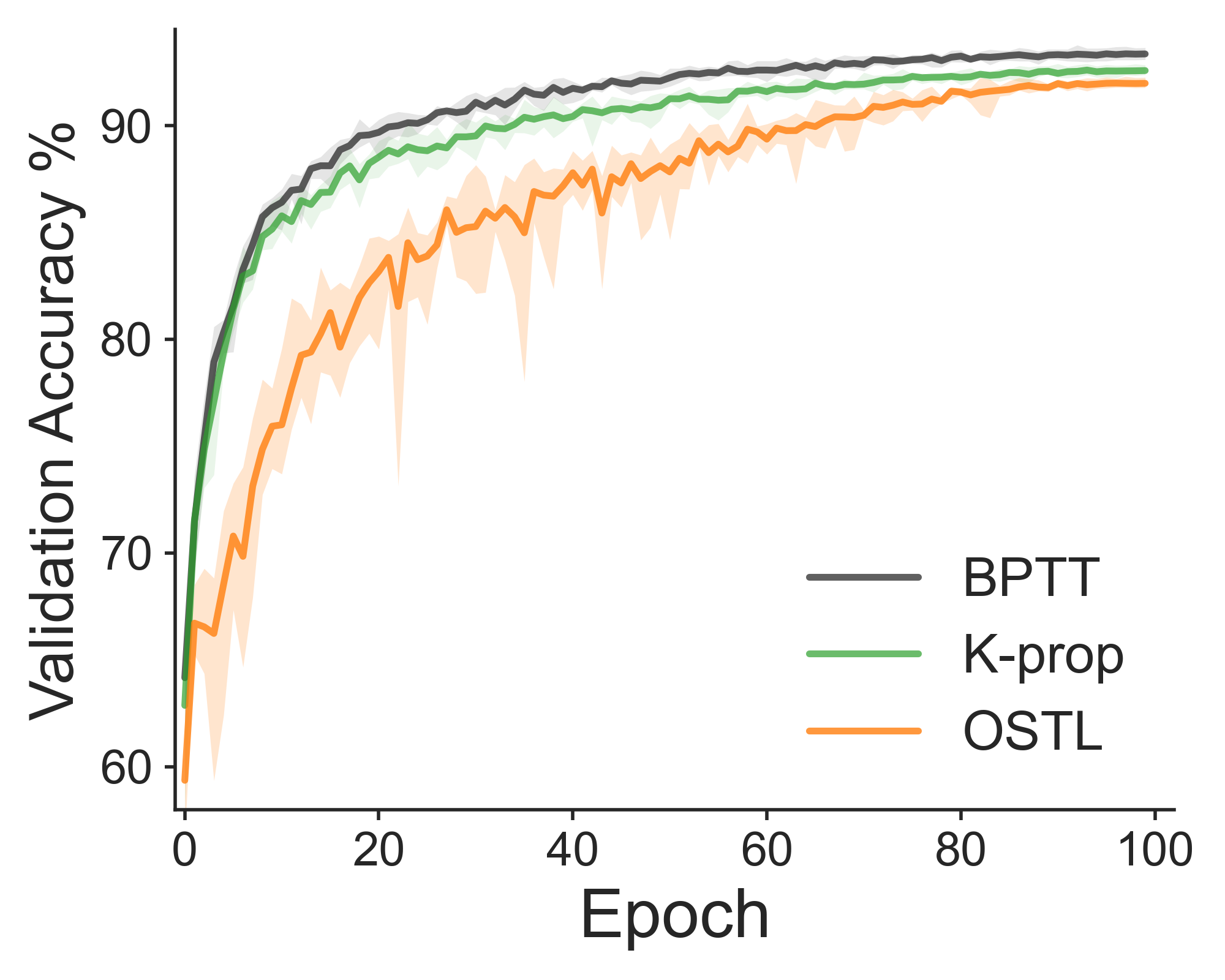}
                \put(3, 72) {\textbf{a}}
            \end{overpic}
            \begin{overpic}[width=0.32\linewidth]{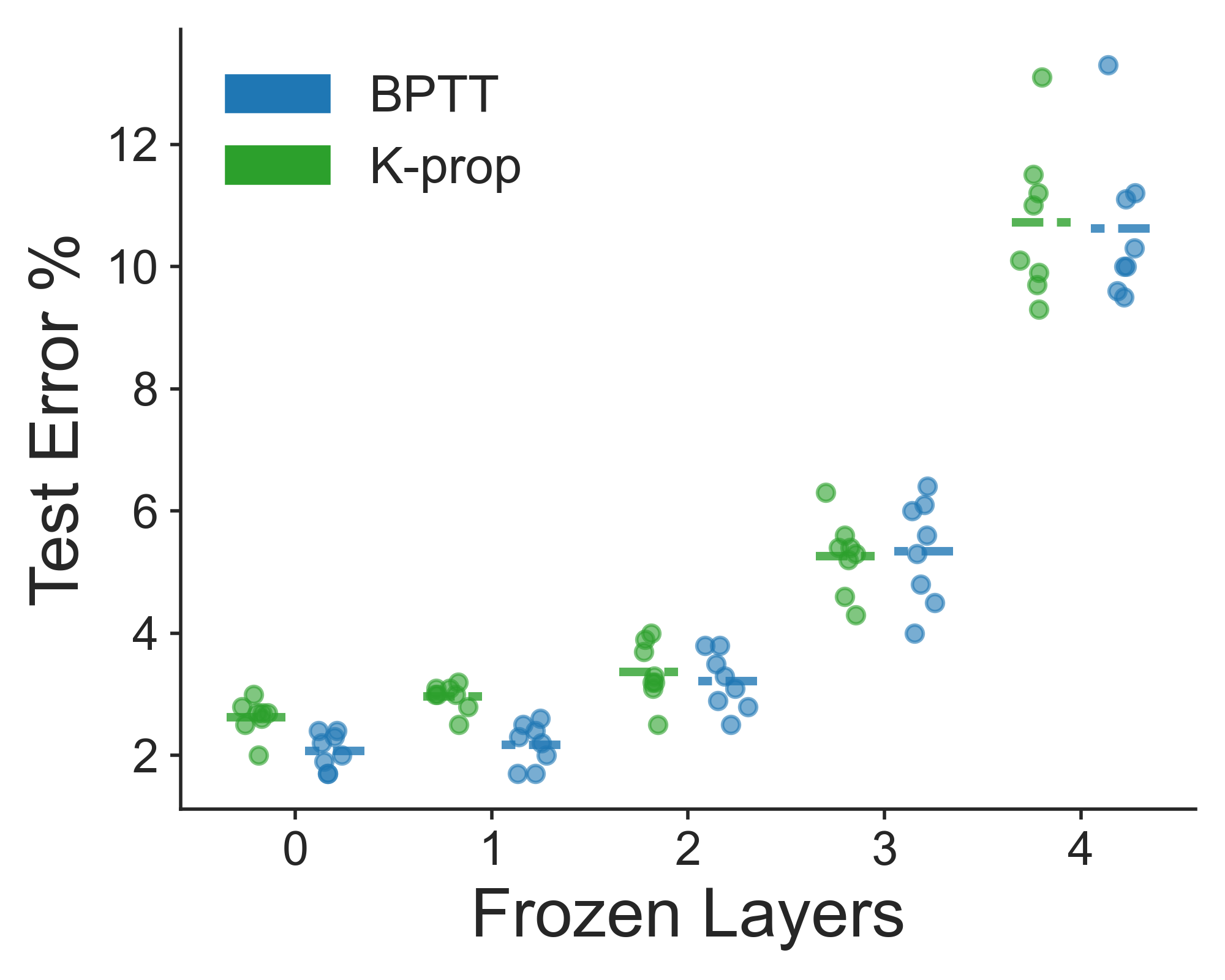}
                \put(3, 72) {\textbf{b}}
            \end{overpic}
             \begin{overpic}[width=0.32\linewidth]{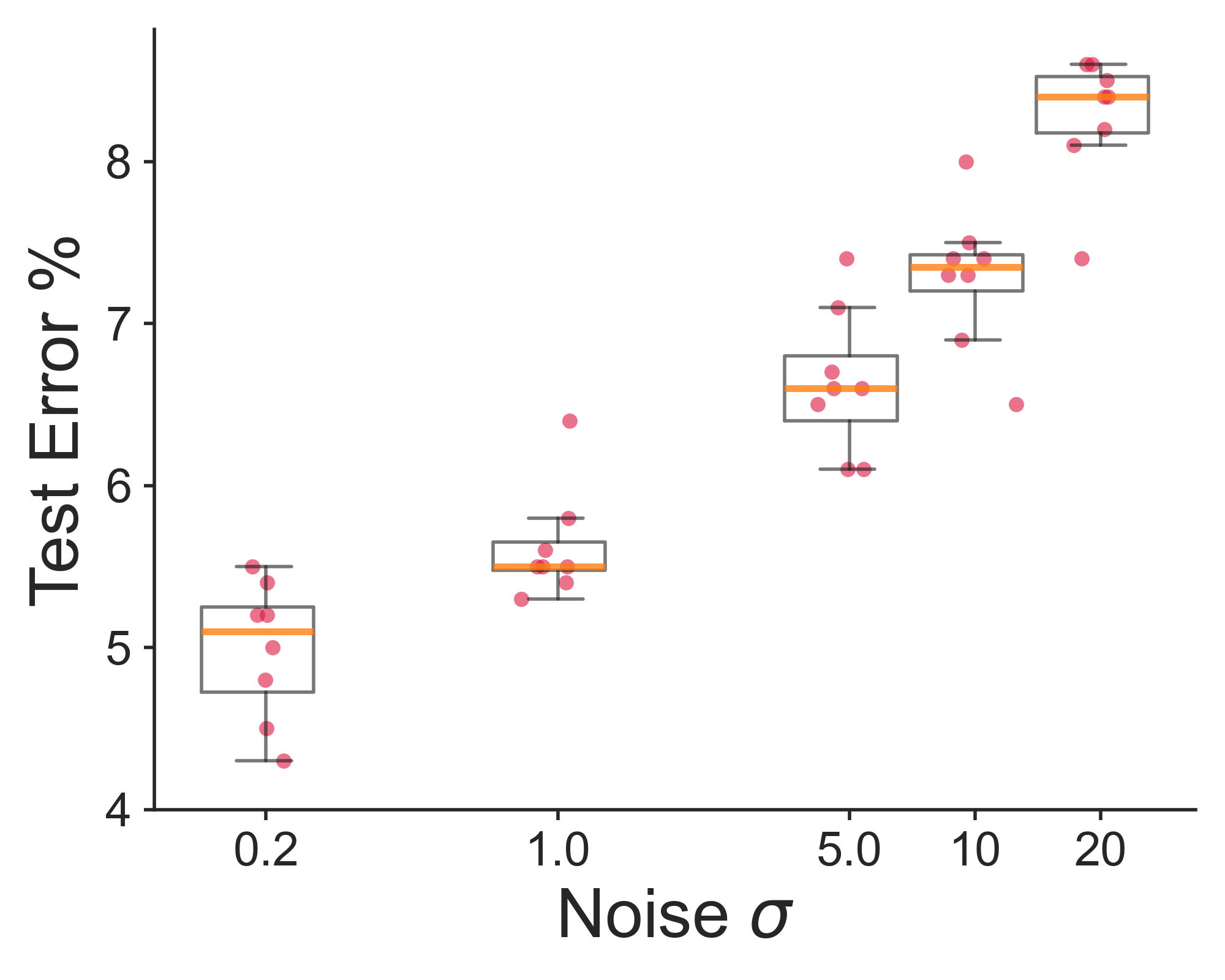}
                \put(3, 72) {\textbf{c}}
            \end{overpic}
            \caption{ Sequential Classification Benchmark learning. 
            (a) Mean, minimum, and maximum validation accuracy during training on GSC \cite{GSC} 20-command task. K-prop outperforms OSTL \cite{OSTL} in final accuracy, convergence speed, and training stability.
            (b) Depth ablation by freezing early layers shows decreasing accuracy with less trained layers. The accuracy loss is similar between K-prop and BPTT, indicating that K-prop provides effective gradients for deep layers. 
            (c) Noise robustness evaluation of the Lk4 in MNIST1d (see \cref{sec:neuromorph} and \ref{sec:noise}). K-prop maintains high accuracy under a broad range of noise levels $\sigma$.
            }
            \label{fig:benchmark}
        \end{figure}

    \subsection{Sequential Classification Tasks Benchmark}
        \label{Sec:MNIST}
        We next verify K-prop in deeper and wider networks on benchmark sequential classification task MNIST1D \citep{MNIST1D}. Prior benchmarks \citep{beebe2026exploring, GLE} report accuracies of approximately 94\% for temporal convolutional networks, 91\% for GRU, and 96\% for human performance. For comparable leaky neural networks trained with GLE \citep{GLE}, performance reaches 91\% with 15K parameters and 93\% with 42K parameters. Using K-prop, we achieve $96.5 \pm 0.5\%$ with a 4-layer leaky network (Lk4, 14K parameters) and $97.5 \pm 0.3\%$ with a 5-layer configuration (Lk5, 21K parameters), outperforming previously reported leaky network baselines while remaining parameter-efficient. 
        Moreover, K-prop closely matches the BPTT loss, whereas OSTL converges to a substantially higher loss (\cref{fig:MNIST1d}).
        
        To evaluate K-prop at a larger scale and on a more realistic dataset, we perform classification on Google Speech Commands (GSC) \citep{GSC}. This task requires substantially longer temporal dependencies to distinguish acoustically similar words such as 'eight' and 'right', making long-range credit assignment critical. On the 20-command task, K-prop achieves an accuracy comparable to that reported for GLE on the 10-command task, despite using approximately half as many parameters (0.55M, \cref{tab:accuracy_comparison}). K-prop also outperforms OSTL in accuracy, convergence speed, and training stability. These results suggest that K-prop more effectively propagates credit across long temporal dependencies than OSTL instantaneous error signals (\cref{fig:benchmark}a).
        
        To assess whether K-prop effectively trains deep layers, we perform a layer ablation study on Lk5 (\cref{fig:benchmark}b). Progressively freezing early layers degrades performance, indicating that K-prop provides meaningful training signals throughout the network depth. 
        Together, these results demonstrate that K-prop scales to deeper architectures and larger, more realistic sequential-learning tasks, supporting its use as an online alternative for training leaky models.

    \subsection{Motor learning through real-time reinforcement learning}
        Motor adaptation in biological systems relies heavily on continuous, online learning, where behavior is adjusted in real time.
        With online learning, we can adapt faster to new environments and make efficient explorations based on learning.
        Such scenarios often provide observations that fall into the regime of a partially observable Markov decision process \citep{kaelbling1998planning}, requiring agents to construct memory for internal state representations.

        To emulate this setting, we use the CartPole and LunarLander tasks from OpenAI Gymnasium \citep{OpenAIgym}. We partially mask the observations by removing both positional variables in CartPole and the horizontal position in LunarLander, which makes the instantaneous observation insufficient for optimal control.
        Although the missing positions can, in principle, be estimated by integrating the observed velocities over time, the agents need to collect historical information correctly without any supervision signal for reconstructing hidden variables. The temporal state inference thus becomes an implicit prerequisite of the reward-maximization task.   

        We employ an online actor–critic framework similar to \citep{eprop}, using TD($\lambda$) for value estimation. Both actor and critic are implemented as leaky neural networks trained with K-prop. Across multiple runs, the agents consistently succeed in reward maximizing, demonstrating that K-prop supports stable policy and value function learning under partial observability of the system and online updates.
        Unlike conventional episodic training \citep{eprop}, our setting emphasizes online adaptation during ongoing interaction without batched updates across episodes. In CartPole, in particular, the weights are updated every $3\mathrm{s}$ of interaction. These results show that K-prop provides effective online optimization beyond supervised learning.

        \begin{figure}[tbp]
            \centering
            \begin{overpic}[width=0.4\textwidth]{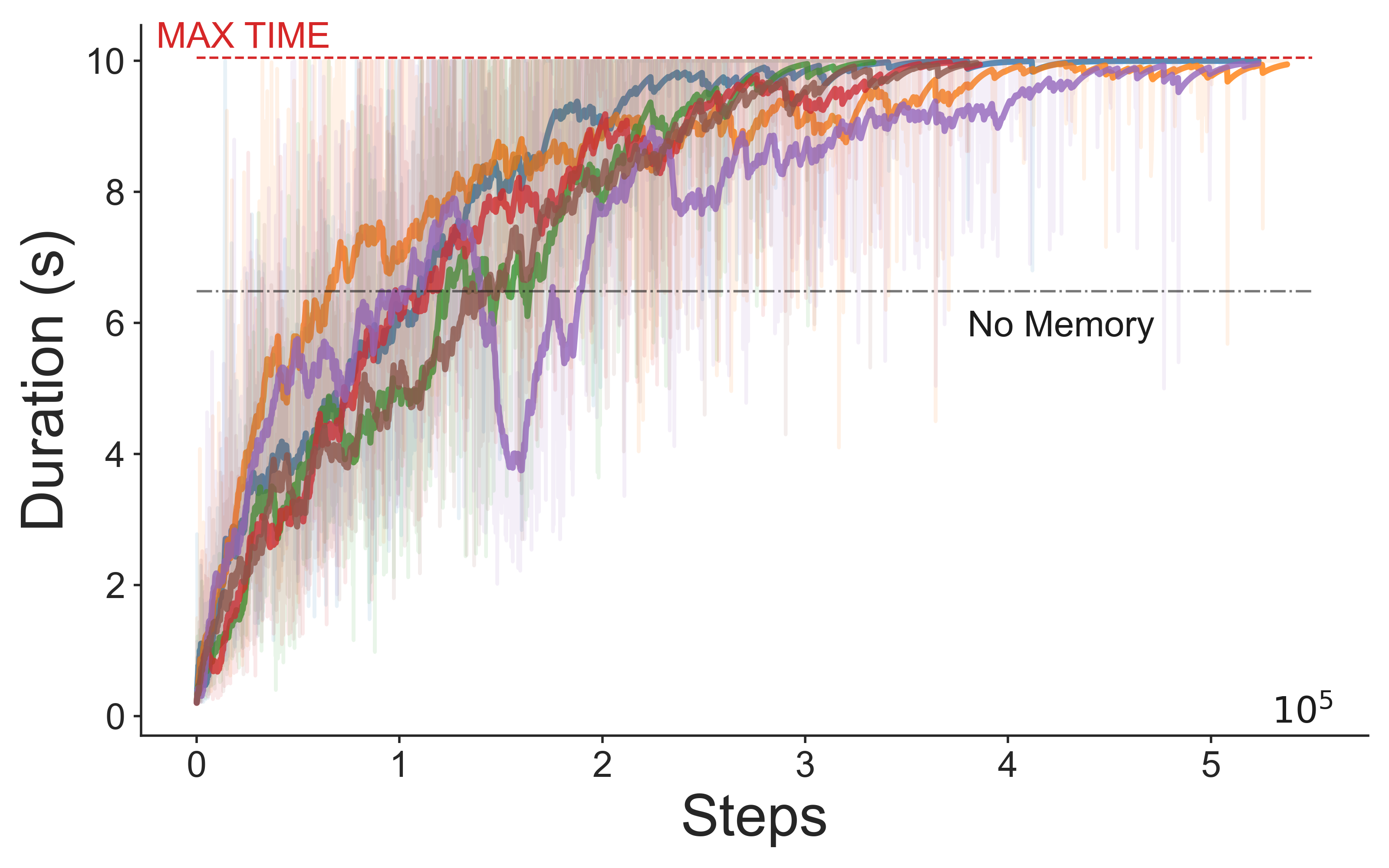}
                \put(3, 57) {\textbf{a}} 
            \end{overpic}
            \begin{overpic}[width=0.17\textwidth, trim=6.7cm 0cm 5.3cm 0cm,clip]{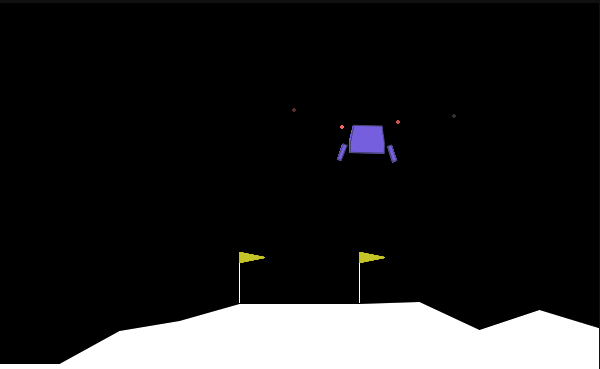}
                \put(-8, 94) {\textbf{b}} 
            \end{overpic}
            \begin{overpic}[width=0.4\textwidth]{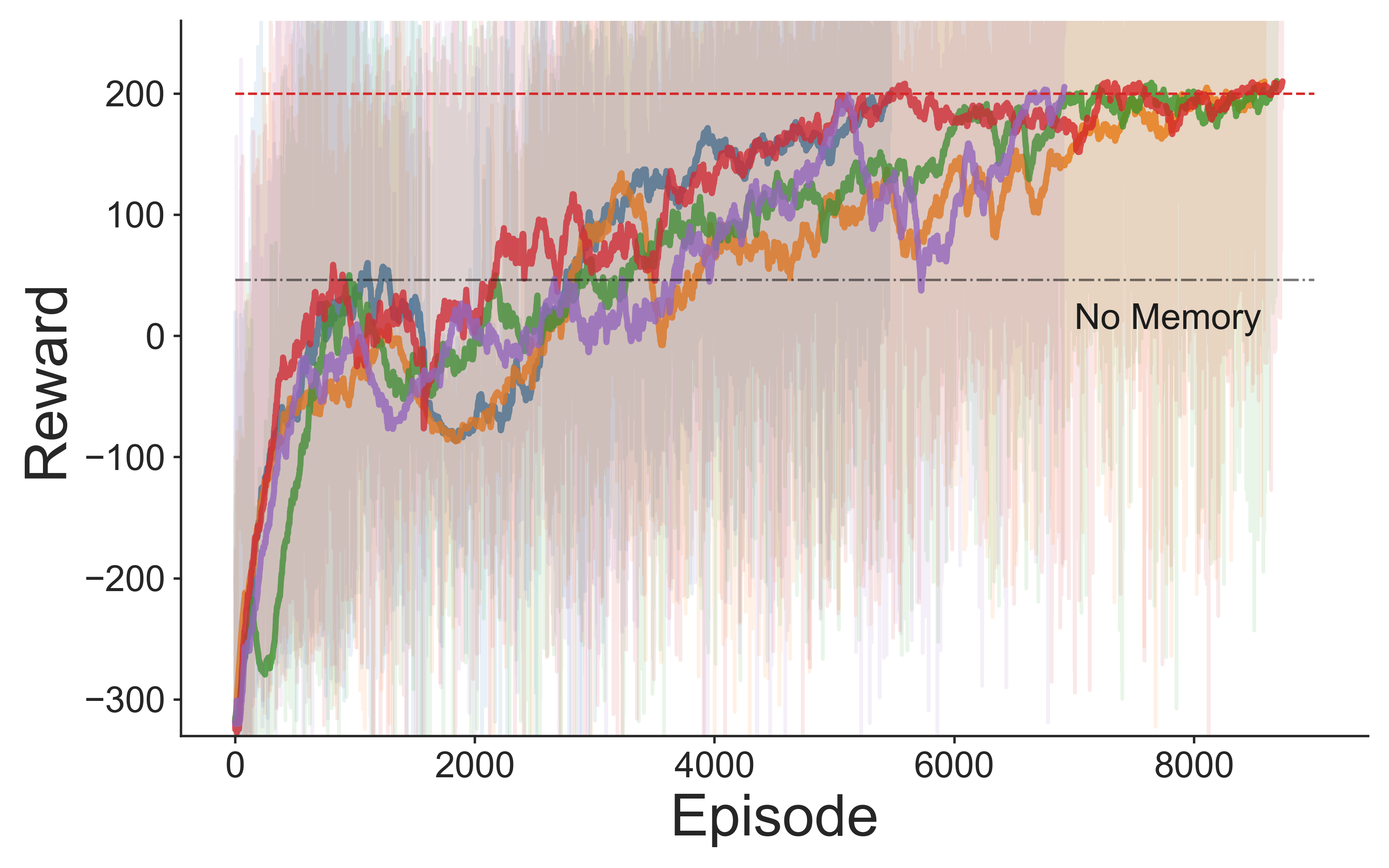}
                \put(5, 57) {\textbf{c}}
            \end{overpic}
            \caption{Motor-skill acquisition through online reinforcement learning. Learning curves for (a) CartPole and (c) LunarLander.
            Faded traces in the background represent individual runs, while solid curves show the corresponding smoothed learning trajectories. Red dashed lines indicate the task-specific success thresholds, and black dashed lines show the performance of an instantaneous, memoryless network. (b) Animation of LunarLander.}
            \label{fig:Cartpole}
        \end{figure}

\section{Discussion}

    \subsection{Computational and memory efficiency}
        K-prop significantly reduces both memory and computation cost compared to RTRL, whose memory cost scales with the number of neurons $n$ as $\mathcal{O}(n^3)$ and computational cost is $\mathcal{O}(n^4)$.
        K-prop maintains deep eligibility traces $\boldsymbol{\zeta}$ with size $\mathcal{O}(n_\tau n^2)$.
        In practice, we divide neurons within each layer into "fast" and "slow" groups (\cref{table:networktau}), leading to $n_\tau = 2n_{\text {layers}}$. Since $n_\tau \ll n$, this yields a substantial reduction in memory requirement.

        The computational cost of K-prop consists of two parts: (i) updating local eligibility traces $\boldsymbol{\zeta}$ and (ii) propagating $\boldsymbol{\mathcal K}$. Both operations scale as $\mathcal{O}(n_\tau n^2)$ per time step, resulting in an overall complexity of the same order. Except for GLE with an extremely efficient memory footprint of $\mathcal{O}(n)$, approximation methods such as RFLO, e-prop, UORO, and OSTL achieve $\mathcal{O}(n^2)$ complexity in both memory and computation (\cref{fig:Cost}).
        K-prop introduces only a modest additional factor $n_\tau$, while providing a more accurate gradient that translates into improved empirical performance.

    \subsection{Robustness against noise in learning}
\label{sec:neuromorph}
     Because K-prop implements low-pass filters at all dynamical stages, both for neurons and synaptic eligibility traces, it has a built-in protection against temporal noise. This is essential for physical networks, both biological and artificial (neuromorphic), where such noise is near-ubiquitous. We evaluate learning under noise by adding Gaussian noise to $\boldsymbol{\mathcal{Z}}$.
     K-prop retains an average MNIST1D accuracy of 94.5\% at noise variance $\sigma=1$ and 92.3\% at $\sigma=10$ (\cref{fig:benchmark}), which are strong noise levels relative to $\boldsymbol{\mathcal{Z}}$ (bounded within $[-1,1]$).
        These results demonstrate that K-prop remains robust even under substantial perturbations.

    \subsection{Weight updating during online learning}
        \label{sec:weightupdate}

        In practice, we observe that applying weight updates at discrete intervals, rather than continuously at every time step, is important for stable learning.
        This observation reflects several fundamental challenges in online gradient-based learning, including exact RTRL.

        First, online gradients implicitly assume that past contributions are evaluated using the current weights, whereas the weights evolve over time. This mismatch introduces inconsistencies between the accumulated eligibility traces and the current parameter state. Second, continuously updating weights interferes with the forward dynamics: the network is simultaneously required to maintain a stable temporal memory while its parameters are being modified, which can disrupt the memory dynamics that generate useful network output. Third, instantaneous gradients are inherently noisy, as they depend on transient activity and error signals; aggregating gradients over a temporal window can reduce variance, analogous to batch averaging in standard stochastic gradient descent.

\section{Conclusion}

    We introduced K-prop, a novel online learning algorithm for feedforward networks of leaky neurons. K-prop significantly reduces the computational and memory cost of RTRL while maintaining a close approximation to the exact gradient, as evidenced by strong directional alignment. Empirically, K-prop achieves performance comparable to RTRL and BPTT on temporal tasks such as N-sum and MNIST1D, and further demonstrates its applicability in reinforcement learning through continuous control in the Cartpole balancing task.
    K-prop outperforms other approximation methods, which we attribute to more accurate temporal credit assignment across multiple time scales and layers.

    From an engineering perspective, K-prop serves as a practical surrogate for RTRL, enabling the study and deployment of online gradient learning in deeper and wider networks. By avoiding the prohibitive cost of RTRL and the severe performance degradation of prior approximations, which potentially conceals the intrinsic limitations, it allows a clearer assessment of online gradient learning methodological properties.
    Its layer-wise structure and compatibility with standard architectures and techniques, such as residual connections, convolutions, normalization and drop-out, make it a viable candidate for integrating online temporal learning into modern deep learning pipelines.

    From a neuroscience perspective, K-prop provides insight into how eligibility traces may operate in deeper neural circuits. In particular, it suggests a framework in which multiple eligibility traces coexist at each synapse, each selectively interacting with different error-related signals, with reward signals provided by distinct neuro-modulators.
    Such a mechanism could support more flexible and expressive temporal credit assignment as compared to simple three-factor learning rules \cite{threefactor}. Meanwhile, K-prop can be extended to other neuron models
    that include, for example, multiple dendritic compartments or adaptive mechanisms by adding structured auxiliary kernels.
    
    Finally, our analysis highlights a key direction for future improvement: the approximation of the activation derivative through $\hat{\rho'}$. The accuracy of K-prop critically depends on the assumption that $\rho'$ varies slowly in time. Relaxing this assumption, especially in regimes with discontinuous or highly non-smooth activations, remains an open challenge. In particular, extending K-prop to spiking neural networks, where $\rho'$ is inherently impulsive, will require more refined surrogate formulations. Addressing this limitation could further broaden the applicability of K-prop to a wider class of temporal and biologically grounded models.

\section*{Acknowledgements}
    We gratefully acknowledge funding from the European Union under grant agreement \#101147319 (EBRAINS 2.0). WS and BL  acknowledge the Swiss National Science Foundation (SNF) for Project funding (Project \#10004971). We would like to express particular gratitude for the ongoing support from the Manfred Stärk Foundation.

\section*{Competing interests}

The authors declare there are no competing interests.

\bibliographystyle{plainnat}
\bibliography{references.bib}

\newpage

\appendix
\section{Mathematical proofs}
    \subsection*{Hierarchical exponential kernels are equivalent to linear combinations of kernels}
    \label{sec:lemmaproof}
        Before analyzing gradients in the network, we first derive a fundamental lemma for this theory.  Hierarchical exponential kernels can be expressed as a linear summation of exponential kernels. This result can be derived using Laplace transforms;
        however, here we provide a derivation based on elementary calculus. Consider a double exponential kernel over a temporal variable $x$:
        \begin{align}
            \overline{ \overline{ x}^{\tau_i}}^{\tau_j} (t) &= \int^t_{t_0} dt'\int^{t'}_{t_0}dt'' \left(\frac1{\tau_j} e^{\frac{t'-t}{\tau_j}} 
            \frac1{\tau_i} e^{\frac{t''-t'}{\tau_i}} x(t'')\right)
            \\
            &= \int^t_{t_0} dt''x(t'')\int^{t}_{t''}dt' 
            \left(\frac1{\tau_j} e^{\frac{t'-t}{\tau_j}}
            \frac1{\tau_i} e^{\frac{t''-t'}{\tau_i}} \right)
            \\
            &= \int^t_{t_0} dt''x(t'')\frac1{\tau_i \tau_j}e^{\frac{t''}{\tau_i}-\frac{t}{\tau_j}}
            \int^t_{t''} dt' e^{\frac{t'}{\tau_j} - \frac{t'}{\tau_i}} 
            \label{eq:IntegralOrder2exponential kernel}
            \\
            &= \int^t_{t_0} dt''x(t'')\frac{1}{\tau_j -\tau_i}
            \left( e^{\frac{t''-t}{\tau_j}}-e^{\frac{t''-t}{\tau_i}}\right)
            \\
            &= \frac{\tau_i}{\tau_i -\tau_j}\overline{x}^{\tau_i} + \frac{\tau_j}{\tau_j -\tau_i}\overline{x}^{\tau_j} \, .
            \label{eq:2exponential kernel}
        \end{align}
        This rule can be applied iteratively to additional exponential kernels. For example, for three exponential kernels:
        \begin{align}
            \overline{\overline{ \overline{x}^{\tau_i}}^{\tau_j}}^{\tau_k} (t) &=  \frac{\tau_i}{\tau_i -\tau_j}\overline{\overline{ x}^{\tau_i}}^{\tau_k} + \frac{\tau_j}{\tau_j -\tau_i}\overline{ \overline{ x}^{\tau_j}}^{\tau_k}\\
            &=\frac{\tau_i}{\tau_i -\tau_j} \left(\frac{\tau_i}{\tau_i -\tau_k}\overline{x}^{\tau_i} +  
            \frac{\tau_k}{\tau_k -\tau_i}\overline{x}^{\tau_k} \right) 
            + \frac{\tau_j}{\tau_j -\tau_j} \left(\frac{\tau_k}{\tau_k -\tau_i}\overline{x}^{\tau_k} +  \frac{\tau_j}{\tau_j -\tau_k}\overline{x}^{\tau_j}  \right) \\
            &=\frac{\tau_i}{\tau_i -\tau_j} \frac{\tau_i}{\tau_i -\tau_k}\overline{x}^{\tau_i} 
            + \frac{\tau_j}{\tau_j -\tau_i} \frac{\tau_j}{\tau_j -\tau_k}\overline{x}^{\tau_j}
            + \frac{\tau_k}{\tau_k -\tau_i} \frac{\tau_k}{\tau_k -\tau_j}\overline{x}^{\tau_k} \, .
        \end{align}
        Then, this lemma can be generalized to an arbitrary number of exponential kernels (a set of distinct time constants $\Upsilon$), where $\mathcal{T}^{\tau_i}_j=1$ when $i = j$, which allows a more compact representation:
        \begin{equation}
           \bm \kappa(x, \Upsilon, t) =
           \sum_{\tau_i\in {\Upsilon}} 
           c_{i}  \, \overline{x}^{\tau_i}(t)
           \;,
           \; \mbox{ with }
           c_{i}= \prod_{
            \tau_j\in {\Upsilon},j \neq i}
           \frac{\tau_i}{\tau_i -\tau_j}  = \prod_{
            \tau_j\in {\Upsilon}} \mathcal{T}^{\tau_i}_{j}
            \;.
            \label{eq:kappa_path} 
        \end{equation}
        The limitation arises when multiple time constants are identical, in which case the expression reduces to the special case of the alpha function, which cannot be written as a simple linear combination. 
        We therefore assume $\tau_i \neq \tau_j$ for any $i\neq j$.
    
    \subsection*{Exact gradient for 2-leaky-layer networks}

        To begin with, we transform the ODE $\tau \dot{\bm u} = -\bm u + \bm W \bm r$ into its solution form $\bm u(t) = \bm W\overline{\bm r}^\tau(t)$ with the initial condition $u(t_0) = 0$.  We use bold symbols to denote vectors and matrices. Consider a network architecture with two hidden layers (and widths of the input, hidden and output layer denoted by $n_\inp$, $n_1$, $n_2$ and $n_\out$), where neurons have exponential filters with different time constants $\tau$, except for the instantaneous output neurons (formally, we only need that neurons from different layers have different time constants). For the last leaky layer, the gradient for minimizing the objective function can be written as:
        \begin{align}
            \dot{\bm W}_{[2]} (t) &\propto - \frac{d \bm u_{[2]}(t)}{dW_{[2]}} \frac{\partial \bm r_{[2]}(t)}{\partial \bm u_{[2]}(t)}\frac{\partial \C(t)}{\partial \bm r_{[2]}(t)} \\
            &=-\bar{\bm r}_{[1]}(t)  \odot\bm \rho'_{[2]}(t)\frac{\partial \C(t)}{\partial \bm r_{[2]}(t)} \, .
        \end{align}
         Here $\bar{\bm r}_l$ without superscript is a matrix of shape $(n_l \times n_{l+1})$ where each column is filtered by corresponding post-synaptic $\tau$. Since the output layer is instantaneous, we can directly compute the error arriving at the last leaky layer: $\bm \epsilon(t)= -\frac{\partial \C(t)}{\partial \bm r_{[2]}(t)} = -\bm W_{\out}^\T \frac{\partial \C(t)}{\partial \bm r_{o}(t)}$, which can be readily obtained. The exact gradient can then be computed online $\propto\bm z_{[2]}\odot\bm \epsilon $.  This reduces to RFLO \cite{RFLO} and E-prop \cite{eprop} for the last hidden layer, which is in fact equivalent to $\zeta^{\tau=0}_{[2]}$ considering downstream instantaneous neurons. 

        For the deeper layer L1, the simple eligibility traces with local $\tau$ are no longer enough, since the temporal gradient $\frac{\partial \C(t)}{\partial \bm r_{[1]}(t'<t)} \neq 0$. The gradient must integrate over temporal contributions $\frac{\partial \bm u_{[2]}(t)}{\partial \bm r_{[1]}(t'<t)}$, which are scaled by the decay term $\frac1{\tau}e^{\frac{t'-t}{\tau}}$:
        \begin{align}
            \dot{\bm W}_{[1]} (t) &\propto -\int^t_{t_0}\bar{\bm r}_\inp(t') \bm \rho'_{[1]}(t') \frac{\partial \bm u_{[2]}(t)}{\partial \bm r_{[1]}(t')} \frac{\partial \C(t)}{\partial \bm u_{[2]}(t)} \  dt'\\
            &=  \int^t_{t_0}\bar{\bm r}_\inp(t')  \bm \rho'_{[1]}(t') \bm W_{[2]}^\T \left(\frac1{\bm \tau}e^{\frac{t'-t}{\bm \tau}} \bm \epsilon(t)\right)   dt' \, .
        \end{align}
        Since we aim to rearrange these terms according to time constants, it becomes difficult to proceed directly with matrix operations. To compute this explicitly, we replace matrix operations with summations. We start by focusing on an arbitrary neuron $\star$ in L1, denote its input weight vector by $ \bm W_{\inp,\star} \in \mathbb{R}^{n_\inp}$, the corresponding eligibility trace by $ \bm z_\star \in \mathbb{R}^{n_\inp}$ and the weight to neuron $i$ in L2 by $W_{i \star}$. We then calculate
        \begin{align}
            \dot{\bm W}_{\inp,\star} (t) &\propto \int^t_{t_0}\bar{\bm r}^{\tau_\star}_{\inp}(t') \rho'_\star(t') \sum_i  W_{i \star}\frac1{\tau_i}e^{\frac{t'-t}{\tau_i}}\epsilon_i(t)  dt'\\
            &= \sum_i\epsilon_i(t)\int^t_{t_0}\bm z_\star(t') \rho'_\star(t')  W_{i \star} \frac1{\tau_i}e^{\frac{t'-t}{\tau_i}}  dt' \label{eq:RTRL}\\ 
            &=  \sum_i W_{i \star} \epsilon_i(t)\overline{\bm z_\star }^{\tau_i}(t) \, . \label{eq:L2}
        \end{align}
        \autoref{eq:RTRL} can be considered as the closed-form solution of the RTRL sensitivity in continuous time, which accounts for the contribution of parameters in $\bm W_\inp$ to each L2 neuron where the error arrives. 
        We next group L2 neurons that share the same time constant and collect all distinct $\tau$'s into the set $\Upsilon_{L2}$. This allows us to merge the coefficients within each group as the kernel gain $\mathcal{K}$ and rewrite the expression as:
        \begin{align}
            \dot{\bm W}_{\inp,\star}(t) &\propto  \sum_{\Upsilon_{L2}} \mathcal{K}^\tau_{\star} \bm \zeta_\star^{\tau}  \;,\;\; \mathcal{K}^\tau_{\star} =\sum_{i| \tau_i = \tau} W_{i \star}\epsilon_i(t) =  \sum_{i} \epsilon_i(t)W_{i \star}\mathcal{T}^\tau_{i}  \, ,\label{eq:L2grouped}
        \end{align}
        where $\mathcal{T}$ compactly replaces the selection condition $\tau_i = \tau$. Therefore, we construct a specialized version of K-prop in which the exact gradient can be computed in two layers networks more efficiently than RTRL.
    
    \subsection*{To three and more leaky layers}
        In the same manner, we can compute the gradient for networks with more leaky layers. To maintain consistent layer indexing for the back-propagation process, we use $[-\text{index}]$ to denote layers by their distance from the output, and indices ${i, j, k, \ldots}$ follow this order, with $\epsilon$ still as the error signal arrives at the last leaky layer. We write the chain rule in an order that facilitates the later transformation into summation form:
        \begin{align}
            \dot{\bm W}_{[\nega3], \star}(t) &\propto - \frac{\partial \C(t)}{\partial u_{[\nega1]}(t)} 
            \int^t_{t_0}  dt'\left(\frac{\partial u_{[\nega1]}(t)}{\partial u_{[\nega2]}(t')}
            \int^{t'}_{t_0}dt''  \frac{\partial u_{[\nega2]}(t')}{\partial r_{[\nega3]}(t'')}\bm z_\star(t'')\right)
            \\
            &= \sum_{i} \epsilon_i (t) \rho'_i(t)
            \int^t_{t_0} dt'\sum_j W_{ij}\frac1{\tau_i} e^{\frac{t'-t}{\tau_i}} \rho'_j(t') 
            \int^{t'}_{t_0}dt'' W_{j\star}\frac1{\tau_j} e^{\frac{t''-t'}{\tau_j}} \bm z_\star(t'')
            \\ 
            &= \sum_i \sum_j W_{ij}  W_{j\star} \epsilon_i (t) \rho'_i(t)
            \int^t_{t_0} dt' \left(\frac1{\tau_i} e^{\frac{t'-t}{\tau_i}} \rho'_j(t') 
            \int^{t'}_{t_0}dt''\frac1{\tau_j} e^{\frac{t''-t'}{\tau_j}}\bm z_\star(t'')\right) 
            \\
            &= \sum_i \sum_j W_{ij}  W_{j\star} \epsilon_i (t) \rho'_i(t)
            \overline{ \rho'_j \overline{ \bm z_\star(t'')}^{\tau_j}}^{\tau_i} (t) \, .
        \end{align}
        To reduce this triple exponential kernel to a double exponential form, we can follow the computation in \autoref{eq:IntegralOrder2exponential kernel}. However, in this case, we encounter an intractable expression due to the temporally varying $\rho'_j(t')$.
        \begin{align}
           \overline{ \rho'_j \overline{\bm z_\star}^{\tau_j}}^{\tau_i} (t) &=   \int^t_{t_0} dt'\frac1{\tau_i} e^{\frac{t'-t}{\tau_i}} \rho'_j(t') 
            \int^{t'}_{t_0}dt''\frac1{\tau_j} e^{\frac{t''-t'}{\tau_j}} \bm z_\star(t'')
            \\
            &= \int^t_{t_0} dt''\bm z_\star(t'')\frac1{\tau_j \tau_i}e^{\frac{t''}{\tau_j}-\frac{t}{\tau_i}}
            \left[\int^t_{t''} dt'  \rho'_j(t') e^{\frac{t'}{\tau_i} - \frac{t'}{\tau_j}} \right] \, .
        \end{align}
        This expression can be approximately handled by extracting $\rho'_j$ from the integral and introducing a surrogate variable $\hat{\rho'}$, such that $ \frac{\rho'_j(t')}{ \hat{\rho'}_{\!j}(t)} \approx 1$ for recent times $t-t' < T\epsilon$. This approximation effectively requires $\rho'$ to vary slowly over time, which is better satisfied by smooth activation functions.
        \begin{align}
           \overline{ \rho_j \overline{ \bm z_\star}^{\tau_j}}^{\tau_i} (t) &=   
            \hat{\rho'_j}(t)\overline{  \frac{\rho'_j}{ \hat{\rho'_j}(t)}\overline{ \bm z_\star}^{\tau_j}}^{\tau_i} (t)\\
            &\approx\hat{\rho'_j}(t)\overline{  \overline{ \bm z_\star}^{\tau_j}}^{\tau_i} (t) \, .
        \end{align}
        Then, according to \autoref{eq:2exponential kernel}, we can convert the double exponential kernel into a summation:
        \begin{align}
            \dot{\bm W}_{[\nega3], \star}(t) &\propto\sum_i \sum_j W_{ij}  W_{j\star} \epsilon_i (t) 
            \rho'_i(t)\hat{\rho'_j}(t)
            \left( \frac{\tau_j}{\tau_j -\tau_i}
            \overline{ \bm z_\star}^{\tau_j}(t)
            +  \frac{\tau_i}{\tau_i -\tau_j}
            \overline{\bm z_\star}^{\tau_i}\right)
            \\
            & = \sum_i \sum_j W_{ij}  W_{j\star} \epsilon_i (t) 
            \rho'_i(t)\hat{\rho'_j}(t)
            \left( \mathcal{T}^{\tau_j}_{i}\bm \zeta_\star^{\tau_j}(t)
            +  \mathcal{T}^{\tau_i}_{j}\bm \zeta_\star^{\tau_i}(t)\right) \, .
            \label{eq:L3grad}
        \end{align}
        In the same sense, we can derive a similar expression for the $\nega4$ hidden layer by introducing an additional summation over the $k$-th neuron in the additional layer. 
        \begin{align}
            \dot{\bm W}_{[\nega4], \star}(t) &\propto \sum_i \sum_j \sum_k W_{ij}  W_{jk}  W_{k\star} \epsilon_i \rho'_i(t)
            \overline{\rho'_j\overline{ \rho'_k \overline{ \bm z_\star}^{\tau_k}}^{\tau_j}}^{\tau_i} (t)
            \\ 
            &\approx \sum_{i, j, k} \sum_k W_{ij}  W_{jk}  W_{k\star}\epsilon_i (t)
             \rho'_i\hat{\rho'_j}\hat{\rho'_k}(t)
            \overline{\overline{  \overline{ \bm z_\star }^{\tau_k}}^{\tau_j}}^{\tau_i} (t)
            \\
            &= \sum_{i, j, k}\sum_k W_{ij}  W_{jk}  W_{k\star} \epsilon_i 
            \rho'_i\hat{\rho'_j}\hat{\rho'_k}
            \left( \mathcal{T}^{\tau_k}_{j} \mathcal{T}^{\tau_k}_{i} \bm \zeta_\star ^{\tau_k}
            +  \mathcal{T}^{\tau_j} _{i} \mathcal{T}^{\tau_j }_{k} \bm \zeta_\star ^{\tau_j}
            + \mathcal{T}^{\tau_i}_{j} \mathcal{T}^{\tau_i }_{k} \bm \zeta_\star ^{\tau_i}\right)(t) \, .
        \end{align}
To generalize this result to arbitrary depth, we define a path $p$ as synaptic chain passing from a initial neuron $i_{[l]}$ at layer $l$ to a next neuron $j_{[l+1]}$ at layer $l\!+\!1$ until we end a neuron $v_{[L]}$ in the output layer, $p = (i_{[l]}, j_{[l+1]} ... , v_{[L]})$. The path weight $w_p $ is defined as the product of all synaptic weights, the rate derivatives $\rho'$, and the error $\epsilon_{p}$ entering the path $p$,  
        
        \begin{equation}
            w_p = \epsilon_{p} \prod_{i \in p} W_{p(i+1), p(i)} \; \hat{\rho'}_{\!\!i+1}  
            \label{eq:weightpath}
        \end{equation}
        where $i$ runs through the components of the path $p$ up to the second last.
        
       We next define the set of paths $\PP_\star$ as set of paths that start at neuron $\star$ in a hidden layer $l$. 
       By $\PP_\star ^\tau$ we denote the subset of paths starting at $\star $ and containing along its path a neuron that low-pass filters the input with $\tau$, i.e.~$\tau \in \Upsilon_{p}$. Using these path sets, we can rewrite the hierarchical summation over layers $\sum_{i, j, k}$ as a single summation over paths:
        \begin{align}
            \dot{\bm W}_{\star}  (t) &\propto \sum_{p \in \PP_\star} w_{p} \sum_{\tau \in \Upsilon_p} \prod_{j \in p^{\setminus\star}}{\mathcal{T}^\tau_{j} \bm \zeta_\star ^{\tau}} \, .
            \label{eq:gradoverpath}
        \end{align}
        Thus, we obtain an approximation of the gradient for all neurons in a network of arbitrary depth, expressed in terms of paths. The remaining question is how to compute this efficiently, since the number of paths in $\PP_\star $ grows exponentially 
        $|\PP_\star | = \prod_{l'=l\!+\!1}^L n_{l'}$, with $l'$ running across all layers downstream of neuron $\star$.
        
    \subsection*{Kernel gain by regrouping summation over $\tau$}
        
        The first observation is that a single $\tau$ can participate in multiple paths, and this multiplicity becomes even larger when the same $\tau$ is shared within a layer. This suggests that maximizing the reuse of $\zeta_\star ^{\tau}$ is beneficial. Therefore, we aim to group the coefficients with respect to $\tau$ by changing the order of summation. We define the set  $\Upsilon_\star$ of all downstream time constants and rewrite \cref{eq:gradoverpath} by 
        \begin{align}
            \dot {\bm W}_\star  (t)&\propto \sum_{\tau \in \Upsilon_\star }\sum_{p \in \PP_\star ^\tau} w_p  \prod_{j \in p^{\setminus \star}} \mathcal{T}^\tau_{j} \bm \zeta_\star ^{\tau}
            = \sum_{\tau \in \Upsilon_\star } \K_\star ^\tau \bm \zeta_\star ^{\tau} 
            \\
            &\mbox{with } \K_\star ^\tau = \sum_{p \in \PP_\star ^\tau} w_p  \prod_{j \in p^{\setminus \star}} \mathcal{T}^\tau_{j} \, .
        \end{align}
        Three challenges arise when computing $\K$: 1. The number of paths in $\PP_\star ^\tau$ grows exponentially with increasing depth. 2. Using $w_p$ requires neurons to access distant downstream information, such as feedback error, synaptic weights, and activations. 3. Neurons must be aware of the precise downstream network structure, i.e., which neurons belong to a given $p^{\setminus \star}$. Since $\zeta$ can be tracked locally in an online manner, it is crucial to determine how $\K$ can be computed efficiently.
        
    \subsection*{Layer-wise propagation of $\K$}

        To address the above challenges, we note that $\K$ can be propagated from layer $l+1$ to $l$, and that this leads to an efficient computational scheme. In a feedforward architecture, each hidden neuron inherits and extends all paths from layer $l\!+\!1$. Define $p_+ = p^{\setminus \star}$ as the path obtained from $p$ by removing neuron $\star$  in layer $l$ and instead starting from neuron $\star\star$ in layer $l+1$. Then we have $ w_p = W_{\star\star, \star} \; \hat{\rho}_{\star \star }' \, w_{p_+}$ and 
        \begin{align}
            \K_\star ^\tau &= \sum_{p \in \PP_\star ^\tau} w_p \prod_{j \in p^{\setminus \star}} \mathcal{T}^\tau_{j}
            \\
            &= \sum_{p \in \PP_\star ^\tau} \mathcal{T}^\tau_{\star \star } \hat{\rho'}_{\!\!\star \star } W_{\star\star, \star }\left(w_{p'} \prod_{j \in p_+^{\setminus \star\star}} \mathcal{T}^\tau_{j}\right) \, .
        \end{align}
        Since many paths share the same node in layer $l+1$, we can rearrange the summation to first sum over neuron $i$ in the next layer:
        \begin{align}
            \K_\star ^\tau & = \sum_{i \in l+1} \hat{\rho'_i}\mathcal{T}^{\tau}_{i}W_{i\star } \sum_{p \in \PP_{i, \tau}} w_{p}   \prod_{j \in {p}} \mathcal{T}^{\tau}_{j}
            \\
             & = \sum_{i \in l+1} \hat{\rho'_i}\mathcal{T}^{\tau}_{i}W_{i \star} \;\K_i^{\tau} \, .
        \end{align}
        Note that $\K^\tau_\star $ must be defined for every $\tau$ and every neuron in the network to support this iterative computation. This equation can be written in matrix form as in \autoref{eq:K-propagation}, which highlights its strong similarity to instantaneous gradient backpropagation.


    \subsection*{$\mathcal{T}$ in other network architectures}
        \label{sec:T}
        Under a rigorous definition, $\mathcal{T}^{\tau}_j=0$ when $\tau$ does not occur on any path through neuron $j$. In a fully connected feedforward network, this applies only to time constants associated with other neurons in the same layer as $j$ and distinct from $\tau_j$. Other architectures can also contain time constants that do not occur on any path through $j$ in different situations.

        Residual networks \citep{resnet} provide one example where skip connections can provide pathways with faster temporal responses than the residual pathway. To apply K-prop to a residual network, we propagate each kernel gain $\mathcal{K}^{\tau}$ along both the skip and residual pathways, then sum the contributions at the merge point for each $\tau$. The eligibility traces $\zeta^{\tau}$ are unchanged. Along the skip pathway, set $\mathcal{T}^{\tau}_j=0$ for time constants that occur only along the residual pathway, since they do not occur on any path through a neuron on the skip pathway.

        More generally, consider a direct connection matrix from a source layer $m$ to target layer $n$, with neuron $j$ in layer $m$.
        Kernels contained by any other, more indirect paths between layers $m$ and $n$ (via other intermediate layers) will neither affect nor be affected by kernels along the direct connection from $m$ to $n$, so their contribution to $\mathcal{T}^{\tau}_{j, [mn]}$ should be zero.
        More formally, let $\Upsilon_{[l<m]}$ denote the set of time constants associated with layers whose indices satisfy $l<m$, and let $\Upsilon_{[l\geq n]}$ denote the set associated with layers whose indices satisfy $l\geq n$. The transformation factor for this connection is
         \begin{equation}
                \mathcal{T}^{\tau}_{j, [mn]} = 
                \begin{cases}
                    \frac{\tau}{\tau -\tau_j} & \tau \in \Upsilon_{[l<m]} \text{ or } \tau \in \Upsilon_{[l\geq n] } \,\\
                    1& \tau_j=\tau\\
                    0&  \text{otherwise}
                    .
                \end{cases}
                \label{eq:T_supp}
            \end{equation}
        This allows us to create the transformation matrices $\mathcal{T}$ efficiently before training without having to evaluate all possible paths.

    \subsection*{Identical or zero $\tau$}
            
        A special case arises when identical time constants appear across multiple layers. As discussed in the main text, this leads to singularities in the kernel decomposition. The singularity can be resolved either by grouping identical $\tau$ into a single basis component or by introducing infinitesimal perturbations. The grouping approach suggests replacing $\boldsymbol{\zeta}$ with higher-order kernels built from nested identical $\tau$, such as the alpha function $\frac{t}{\tau}e^{-t/\tau}$. The transformation factor $\mathcal{T}$ and the propagation rule would then need to be modified accordingly to accommodate the new basis. While the grouping approach is theoretically valid, it introduces additional complexity in both the kernel representation and the propagation rules. To prevent grouping, we consider infinitesimal perturbations for simplicity and consistency with the main formulation.

        Biologically, different time constants across layers are consistent with the hierarchical increase of time constants in the primate cortex \citep{murray2014hierarchy}.
        Empirically, we do not observe any benefit from using identical or near-identical time constants across layers. In our experiments, time constants are chosen to be distinct across layers. We further verify that small perturbations (e.g., differences on the order of $0.1$) do not affect numerical stability or performance.

         Another special case is $\tau=0$. This case is taken into account because $\mathcal{T}^{\tau=0}_{\!\! j} = 0$ for any neuron $j$ including $\tau_j =  0$.  An instantaneous pathway with co-existence of $\tau=0$ neurons across layers does not cause any singularity, which collapses into simple backpropagation.

        
\subsection*{Practical implementation of K-prop}

    A few notes for practical implementation: 1. We define $\zeta^{\tau_{\mathrm{up}}} = 0$, so each layer can omit tracking the upstream components of $\bm \zeta$. 2. A special case arises for the first leaky layer: its $\tau$ does not need to be included in $\Upsilon_\net$ (since there are no upstream neurons), and its nonlinearity $\rho_{L1}$ does not affect the approximation. 3. $\K$ propagation is only required when an external error signal is present (otherwise all $\K = 0$). In many tasks, including real world scenarios, error signals are sparse in time, allowing $\K$ propagation to be triggered in an event-based manner. Altogether, in practical implementations, K-prop has both computational and memory costs on the order of half of $\mathcal{O}(n_\tau n^2)$. Under this efficient implementation, we can plot the cost comparison:

    \begin{figure}
        \centering
        \includegraphics[width=0.64\linewidth]{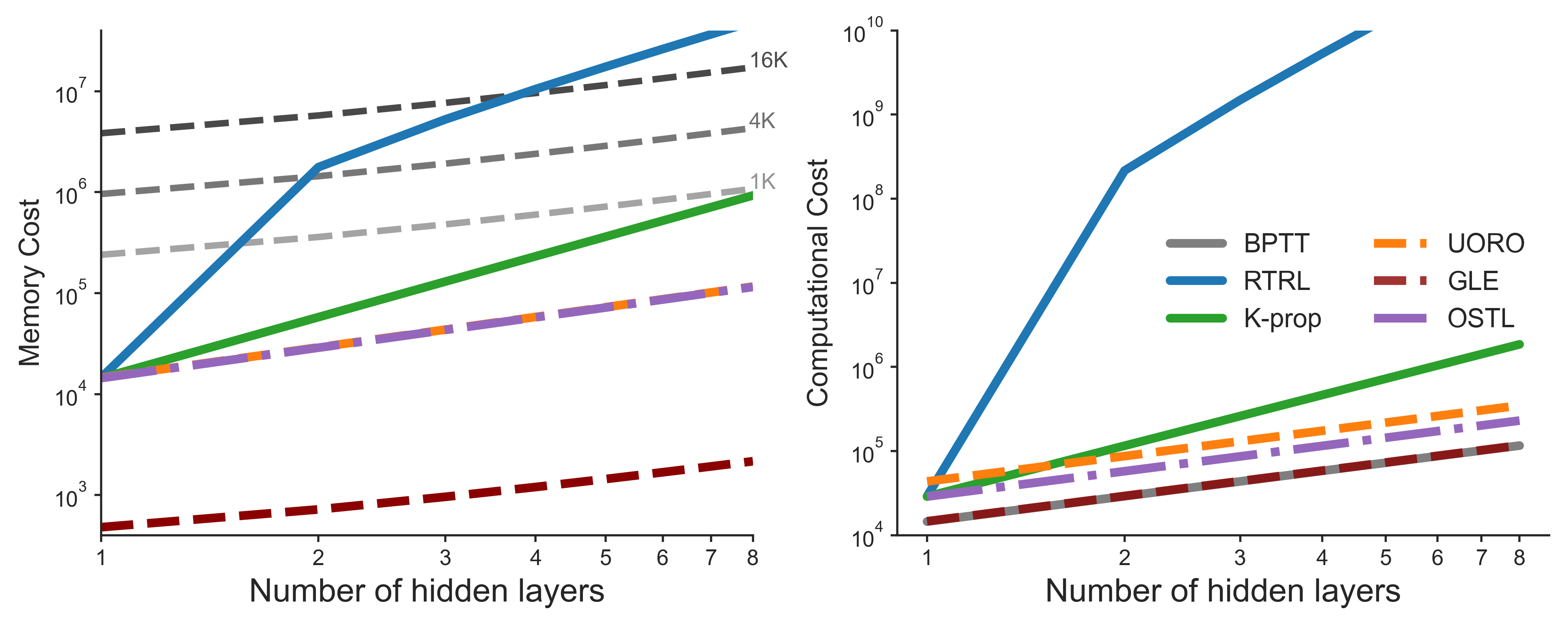}
        \includegraphics[width=0.32\linewidth]{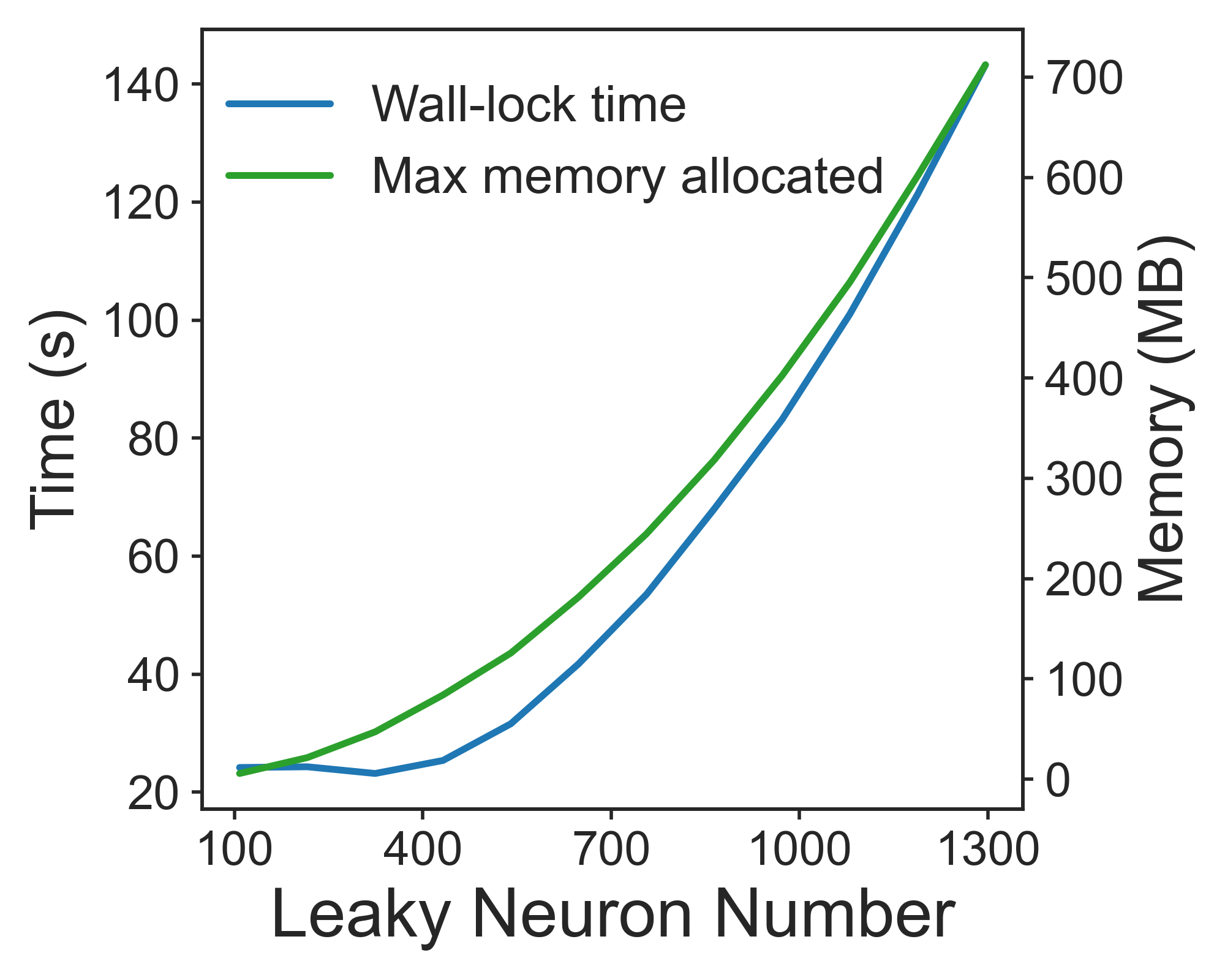}
        \caption{Compare the cost under multi-layer architecture. \textbf{Left}: Memory cost \textbf{Middle}: Computational cost per time step.  GLE has $\mathcal{O}(n)$ memory and $\mathcal{O}(n^2)$ computational cost. All costs are computed in optimal layer-wise implementations instead of directly from scaling law. \textbf{Right}: Wall-clock release time and maximum memory allocation per epoch on the MNIST1D dataset, measured across an increasing number of leaky neurons in 5-layer networks. Experiments were conducted on a local machine equipped with an NVIDIA RTX 4060 Laptop GPU (8 GB). Overall running time reflects additional latency introduced by temporal loop overhead.}
        \label{fig:Cost}
    \end{figure}

    This reduction is well illustrated in the CartPole task, where K-prop reduces RTRL memory cost by $28\times$ and computational cost by $1{,}684\times$, turning an otherwise intractable method into a practical one. Although BPTT is computationally efficient, it incurs a memory cost that scales with sequence length $T$, as it requires storing the full trajectory. 
    In comparison, K-prop reduces memory usage by up to $57\times$, at the expense of a moderate ($\sim 4\times$) increase in computation relative to BPTT, demonstrating a classical space-time trade-off.
    Simulations are programmed in Python with PyTorch. Code is available through \url{https://github.com/Baskin-LIU/Kernel-Propagation#}

    \begin{algorithm}[H]
    \caption{Kernel Propagation (K-prop) for Online Learning in Leaky Networks}
    \label{alg:kp_online}
    \begin{algorithmic}[1]
        \State \textbf{Initialization:} Network with $L$ layers, time constants $\tau$, step size $dt$
        \State Initialize weights $\{\bm W, \bm b\}$ randomly \Comment{bias is treated as weight for $r=1$}
        \State Construct transformation factor matrices $\{\mathcal{\bm T}\}$
    
        \For{each iteration / episode}
            \State Reset $\bar{\bm r}_{[l]} \gets 0$, $\bm \zeta_{[l]} \gets 0$, $\Delta \bm W_{[l]} \gets 0$
    
            \For{each time step $t$}
                \For{layer $l = 1$ to $L$}
                    \Comment{Forward dynamics}
                    \State $\bar{\bm r}_{[l]} \gets (1 - \frac{dt}{\tau_l}) \bar{\bm r}_{[l]} + \frac{dt}{\tau_l} \bm r_{[l]}$
                    \State $\bm u_{[l]} \gets \bm W_{[l]} \bar{\bm r}_{[l-1]} + \bm b_{[l]}$
                    \State $\bm r_{[l]} \gets \rho(\bm u_{[l]})$
                    \State $\bm \zeta_{[l]} \gets (1 - \frac{dt}{\boldsymbol{\tau}_{\mathrm{net}}}) \odot \bm \zeta_{[l]} + \frac{dt}{\boldsymbol{\tau}_{\mathrm{net}}} \; (\rho'_{[l]} \otimes \bar{\bm r}_{[l-1]})$
                    \Comment{Deep eligibility trace update}
                \EndFor
    
                \If{$\frac{\partial \C}{\partial \bm r_L} \neq 0$}
                    \Comment{Check if error signal is available}
                    \State $\bm \K_L^\tau \gets \frac{\partial \C}{\partial \bm r_L} \quad \forall \tau \in \boldsymbol{\Upsilon}_{\mathrm{net}}$
                    \Comment{Initialize $\bm \K$ at output layer}
                    \For{layer $l = L$ down to $1$}
                        \Comment{Backward kernel propagation}
                        \State $\bm \K_{[l-1]}^\tau \gets \bm W_{[l]}^\top \big( \hat{\bm \rho'}_{\![l]} \odot \bm \K_{[l]}^\tau \odot \mathcal{\bm T}_{[l]}^\tau \big)$
                        \State $\dot{\bm W}_{[l]} \gets \sum_{\tau \in \boldsymbol{\Upsilon}^l_{\mathrm{down}}} \mathrm{diag}(\bm \K_{[l]}^\tau)\,\bm \zeta_{[l]}^\tau$
                        \State $\Delta \bm W_{[l]} \gets \Delta \bm W_{[l]} + \dot{\bm W}_{[l]}$
                        \Comment{Gradient accumulation}
                    \EndFor
                \EndIf
    
                \If{Update weight}
                    \For{layer $l = 1$ to $L$}
                        \State $\bm W_{[l]} \gets \bm W_{[l]} - \eta \, \Delta \bm W_{[l]}$
                        \State $\Delta \bm W_{[l]} \gets 0$
                    \EndFor
                \EndIf
            \EndFor
        \EndFor
    \end{algorithmic}
    \end{algorithm}

\section{Experimental details}

    \subsection{MNIST1D  dataset}
        MNIST1D converts digit patterns analogous to the handwritten images in MNIST \citep{MNIST} into one-dimensional temporal sequences. Each sample spans 360ms and contains a digit pattern of approximately 90ms embedded at a random temporal position, requiring both temporal localization and pattern recognition. Apart from increasing the number of time steps, we use the default sample-generation settings of \citet{MNIST1D}, including the default training/test split of 4000/1000 samples.
    
    \subsection{GSC dataset}
        The Google Speech Commands (GSC) dataset \citep{GSC} comprises utterances lasting approximately 0.8-1s, recorded from multiple speakers. We evaluate the 20-class command-label subset, rather than the 12-class subset used in \citet{GLE}. The selected commands are 
        ['zero', 'one', 'two', 'three', 'four', 'five', 'six', 'seven', 'nine', 'eight','yes', 'no', 'up', 'down', 'left', 'right', 'on', 'off', 'stop', 'go'].
        We use the default training, validation, and test splits provided by the PyTorch dataset. The original audio waveforms are converted into 80-channel log-Mel spectrograms and interpolated to 500 time frames, corresponding to a time step of approximately 2ms. We apply data augmentation based on SpecAugment \citep{Specaugment}, including frequency masking over four channels, temporal shifts of up to 25ms, and temporal warping of up to 25ms.
    
        \begin{figure}
            \centering
            \includegraphics[width=0.4\linewidth]{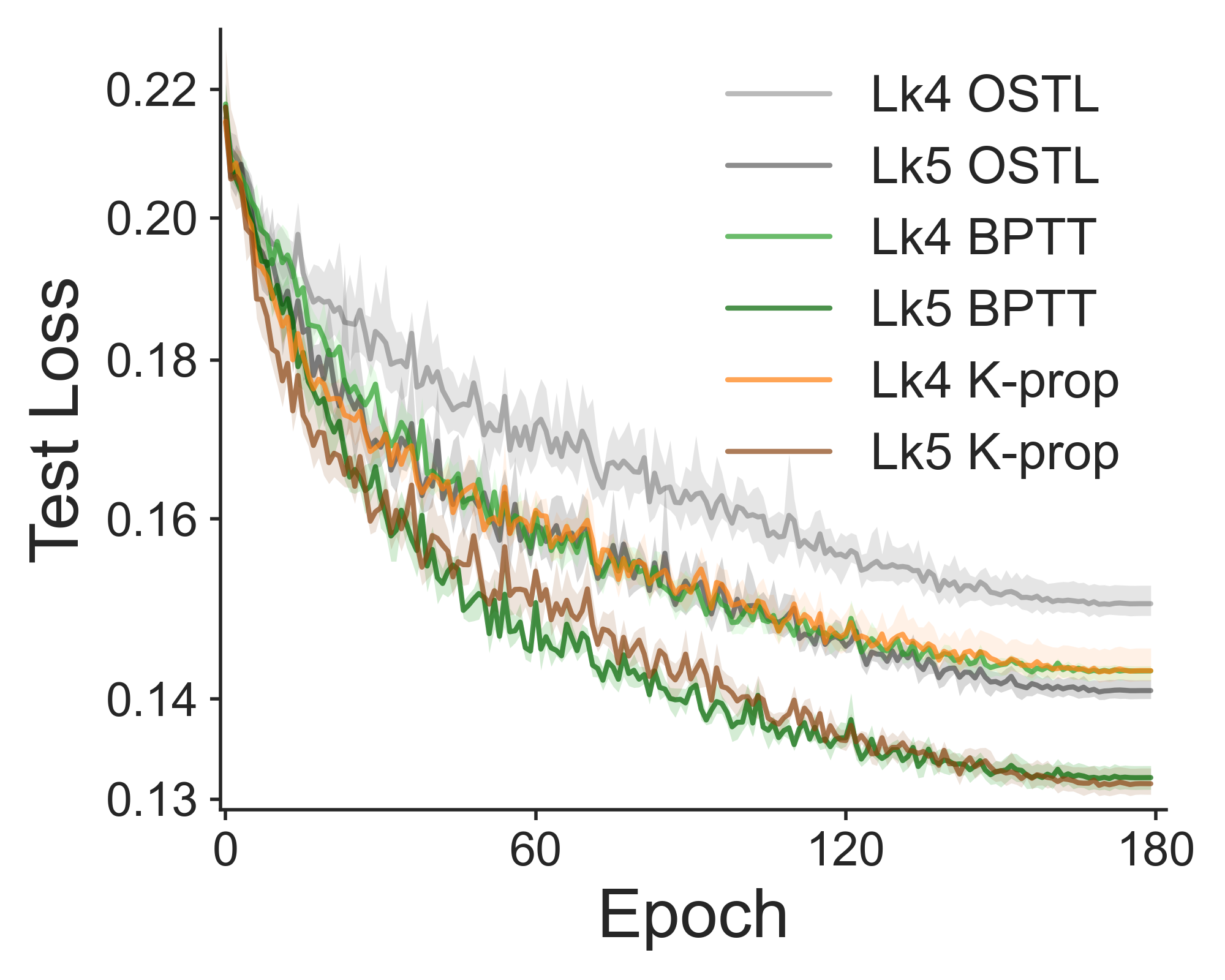}
            \includegraphics[width=0.4\linewidth]{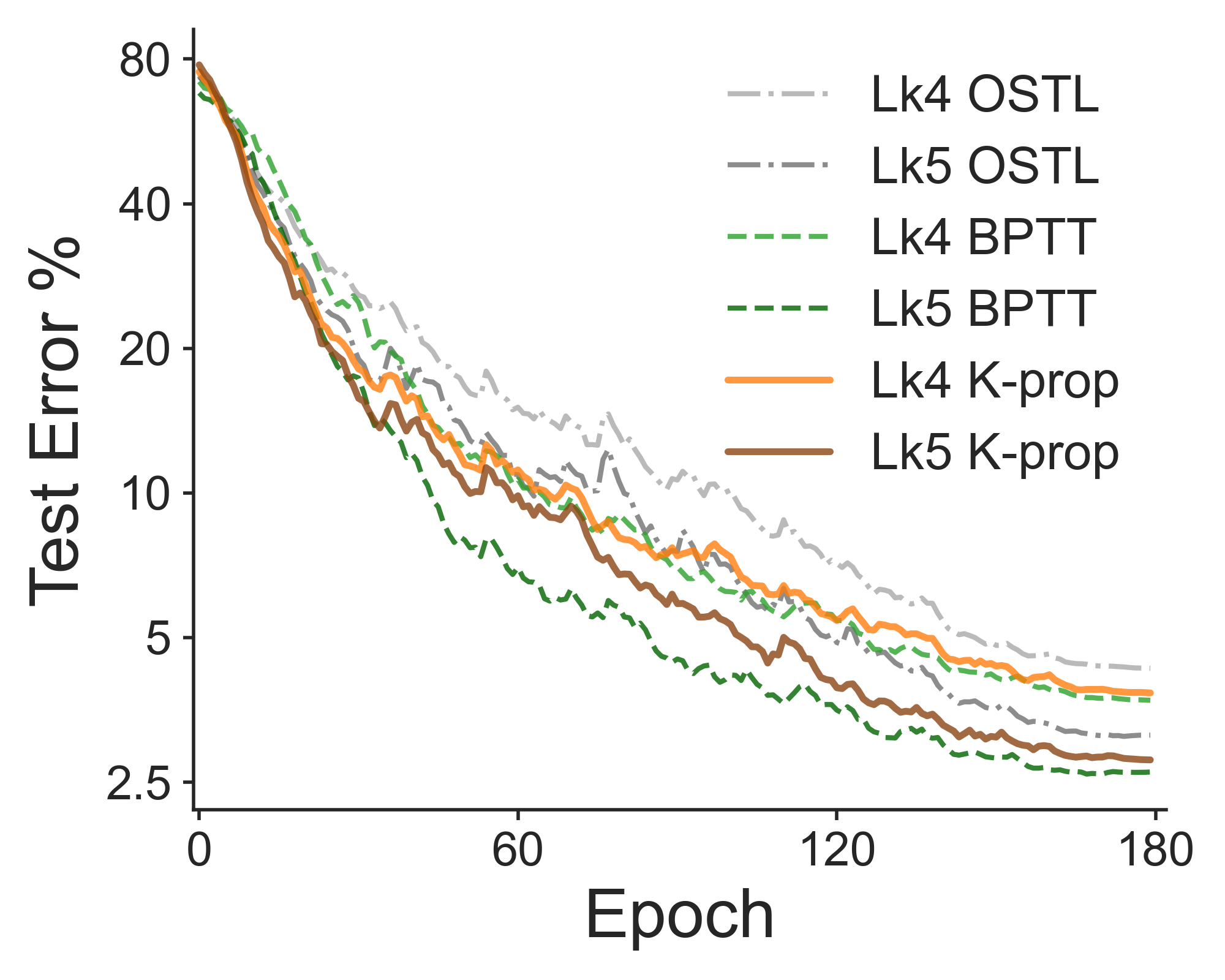}
            \caption{\textbf{Learning curves on MNIST1D.}
                Left: Mean, minimum, and maximum training loss. K-prop closely matches the loss achieved by BPTT, whereas OSTL converges to a substantially higher loss.
                Right: Mean test error for the Lk4 and Lk5 networks. Curves are smoothed for visualization.}
            \label{fig:MNIST1d}
        \end{figure}

    \subsection{Loss and Optimizer}

        In all experiments, we use $r =\mathrm{tanh}(\alpha u)$ with $\alpha \approx 0.5$, making it behave similarly to a centered sigmoid. The learning rate decays exponentially in N-sum and follows a cosine annealing schedule (180 epochs total) in MNIST1D. All methods share the same learning rate scheduler. We use standard gradient descent for N-sum and the Adam optimizer for all other tasks. For Adam, $(\beta_1, \beta_2)$ are set to (0.9, 0.999) for BPTT and (0.8, 0.995) for K-prop to reduce momentum and encourage stronger averaging. We have verified that (0.8, 0.995) on BPTT yields worse performance, excluding the advantage brought by the optimizer setup.
    
        The loss is defined online in all experiments. For N-sum, we use logistic output units with a mean-squared-error loss. For MNIST1D and GSC, cross-entropy loss is evaluated at every time step within a designated response period: the entire sequence for MNIST1D and the final 600,ms for GSC. The total loss is therefore the sum over all answering time steps $T$, $\sum_{t \in T} \mathcal{L}(u_o(t), y)$.
        
        We use a set of random seeds {0, 42, 428, 781, 3407, 115, 256, 666} for training.

\subsection{Reinforcement learning tasks}

    For the Cartpole environment, we extend the termination boundaries to $\pm 36^\circ$ (from $\pm 24^\circ$) for pole angle and $\pm 3.6$ (from $\pm 2.4$) for cart position, allowing longer and more variable episode. To obtain more continuous inputs, observations from each environment step (20 ms) are interpolated by inserting 3 intermediate frames to obtain $dt=5ms$. Actor has two output neurons with softmax activation, producing the probability of Left and Right. The action choice at each step is sampled by a Bernoulli distribution with the possibility.

    In LunarLander, the agent must smoothly land a spacecraft within a target zone using four discrete actions: \texttt{do nothing}, \texttt{fire left engine}, \texttt{fire main engine}, and \texttt{fire right engine}. The environment follows the default OpenAI Gymnasium configuration, except that the horizontal position is masked from the observation. Frame interpolation is not required, and actions are sampled in the same manner as in CartPole. The task is considered successfully learned when the recent average reward exceeds 205.
    
    The reinforcement learning algorithm follows \cite{eprop}, which transforms TD($\lambda$) $\dot{W} \propto \bar{\delta} \frac{\partial V}{\partial W} \to \delta \overline{\frac{\partial V}{\partial W}}$. The critic propagates a scalar value signal, while the actor propagates a cross-entropy loss using the chosen action as the target label. The propagated gradient is first accumulated in an auxiliary term that decays with $\lambda$, and is then scaled by $\delta(t) = R(t) + \gamma  V(t + dt) - V(t)$, where $\gamma=0.99$ is the discount rate. $V(t + dt)$ is computed by a target critic network that is softly updated toward the critic with a factor of $0.05$. We devices a special learning rate scheduler: After each episode, the learning rate is decreased by a factor of $\times 0.997$ if the episodic reward is lower than the running average, and increased by $+0.001$ when episodic reward is larger. otherwise. Due to the inherent instability of actor–critic methods and the absence of batch learning, we clip both $\delta$ and the gradient norm to prevent value explosion. 

\subsection{Implementation configurations}
    \begin{table}[ht]
      \centering
      \begin{tabular}{lllllllll}
        Model    &$dt$ (ms)   &Leaky layers&L1& L2&  L3& L4& L5 &\\
        N-sum    & 0.25       &3& [0, 6.28]& {0.25, 3, 6}& {2, 4, 10}& &  &\\
        Lk4      &1           &4& [0, 10]& {3, 6}& {2, 7}& {2.5, 8}&  &\\
        Lk5      &1           &5& [0, 10]& {3, 6}& {2, 7}& {1, 8}& {2.5, 6.6} &\\
 GSC& 2& 4& [0, 40]& {3,12, 24}& {4, 16, 36}& {50, 400}& &\\
        Cartpole &5          &3& [0, 40]& {5, 10}& {11, 24}& &  &\\
 Lunarlander& 5& 3& [0, 20]& {5, 20}& {6, 10}& & &\\               
      \end{tabular}
      \caption{Network time constants (ms). Since the first layer does not need to be included in $\Upsilon_{\net}$, it is preferable to use a wider range of time constants in the first layer and fewer in downstream layers. Accordingly, we assign the first-layer time constants $\tau$ to be evenly spaced in log space. In the other layers, we use two distinct $\tau$ values per layer, divided into two equal groups. We find that introducing more distinct $\tau$ values within a layer does not yield notable improvements, as long as both a group of “fast” and a group of “slow” neurons are present. Square brackets indicate the log-space range for L1.}
      \label{table:networktau}
    \end{table}
    The issues from online weight updating \cref{sec:weightupdate} can be mitigated either by using very small learning rates or by accumulating gradients over a time interval and applying periodic updates. We find the latter more effective, as it preserves learning speed while improving stability. This design choice is also consistent with biological observations, where long-term synaptic plasticity operates on slower timescales rather than instantaneously \citep{early_late_plas, plasticity_timescale}.

    Meanwhile, our experiments indicates a necessity of "fast" neurons with much shorter $\tau$ than typical neuron membrane potential time constants. Such range of time constants could be achieved through modulation of the membrane conductance \citep{destexhe2003high} or prospective coding \citep{haider2021latent,NLA,brandt2024prospective}.
    
    Both N-sum, Cartpole, and LunarLander simulations can run on CPU hardware within 20 minutes. MNIST1D and GSC benefits from GPU acceleration due to larger matrix operations. A mid-range GPU such as a GTX 4060 reduces MNIST1D training time from approximately 4 hours on CPU to around 1.5 hours. Additional experimental details are provided below.
    
    \begin{table}[h]
      \centering
      \begin{tabular}{llllll}
        Model      & Network width                &  $\alpha$  & Learning rate & Update Period &Batch size\\
        N-sum      & 6-6-6-2& 0.6        & 0.1           & 50 ms &1\\
        Lk4        & 36-54-60-60-72-10& 0.6        & 0.01          & 180 ms &100\\
        Lk5        & 54-60-60-60-60-72-10& 0.5        & 0.01          & 180 ms &100\\
 GSC& 240-360-360-400-400-20& 0.5& 0.001& 1s&256\\
        Cartpole   & 32-40-30-20-2& 0.6        & 0.01          & 3 s &1\\
 Lunarlander& 120-120-80-20-4& 0.6& 0.0002& -&1\\               
      \end{tabular}
      \caption{Network configurations}
      \label{table:networkconfig}
    \end{table}

    \subsection{Learning under noise}
    \label{sec:noise}
    
        To investigate the sensitivity of K-prop to noise in its eligibility dynamics, we inject an independent Gaussian perturbation into each integration step of \cref{eq:xi}, with $\mathfrak{n}\sim\mathcal{N}(0,\sigma^2)$. We evaluate the Lk4 network on MNIST1D using noise levels $\sigma\in{0.2,1.0,5.0,10,20}$. K-prop remains effective across a broad range of noise levels. Given that $z$ is bounded within $[-1,1]$, the results indicate that the perturbations must be large relative to the driving signal before learning performance deteriorates substantially.

        We attribute this robustness to two properties of K-prop:
        
        \begin{enumerate}
        \item Independent Gaussian perturbations are smoothed by the leaky integration dynamics. When driven by white noise, these dynamics produce an Ornstein--Uhlenbeck-like process in which the gradient signal contributes to the drift.
        \item The resulting noisy gradient estimates are further averaged over time through gradient accumulation and the optimizer dynamics.
        \end{enumerate}
        
        These results provide preliminary evidence that K-prop can tolerate the noisy dynamics expected in physical computing substrates.

\section{Robustness of the approximation}

 \subsection{Tuning the non-linearity}
        
    In terms of nonlinearity, the K-prop approximation works better for smoother and flatter functions such as logistic and tanh. Scaling down the input $\rho(\alpha x), 0<\alpha<1$ slows down the variation and thus improves the approximation. However, a smaller $\alpha$ reduces the expressiveness of the nonlinearity, which may degrade performance. Therefore, there is a trade-off between approximation quality and nonlinearity, and a “sweet spot” typically exists that yields the best performance. The sweet spot depends on the network depth and $\tau$ configuration, as illustrated in the MNIST1D experiment.

    In Lk4, the test accuracy increases significantly from $\alpha = 0.2$ and reaches an optimum around $\alpha = 0.5, 0.6$. The performance slightly decreases as $\alpha$ approaches $1$, while still maintaining an accuracy of $94\%$, suggesting that K-prop preserves a usable gradient quality. In Lk5, the sweet spot shifts to $\alpha = 0.5$. We therefore choose $\alpha=0.6$ for Lk4 and $\alpha=0.5$ for Lk5. Meanwhile, models with $\alpha=1$ trained with BPTT do not show significant improvement, indicating that $\alpha$ scaling is an effective hyperparameter that can be tuned for K-prop.

    \begin{figure}
            \centering
            \includegraphics[width=0.32\linewidth]{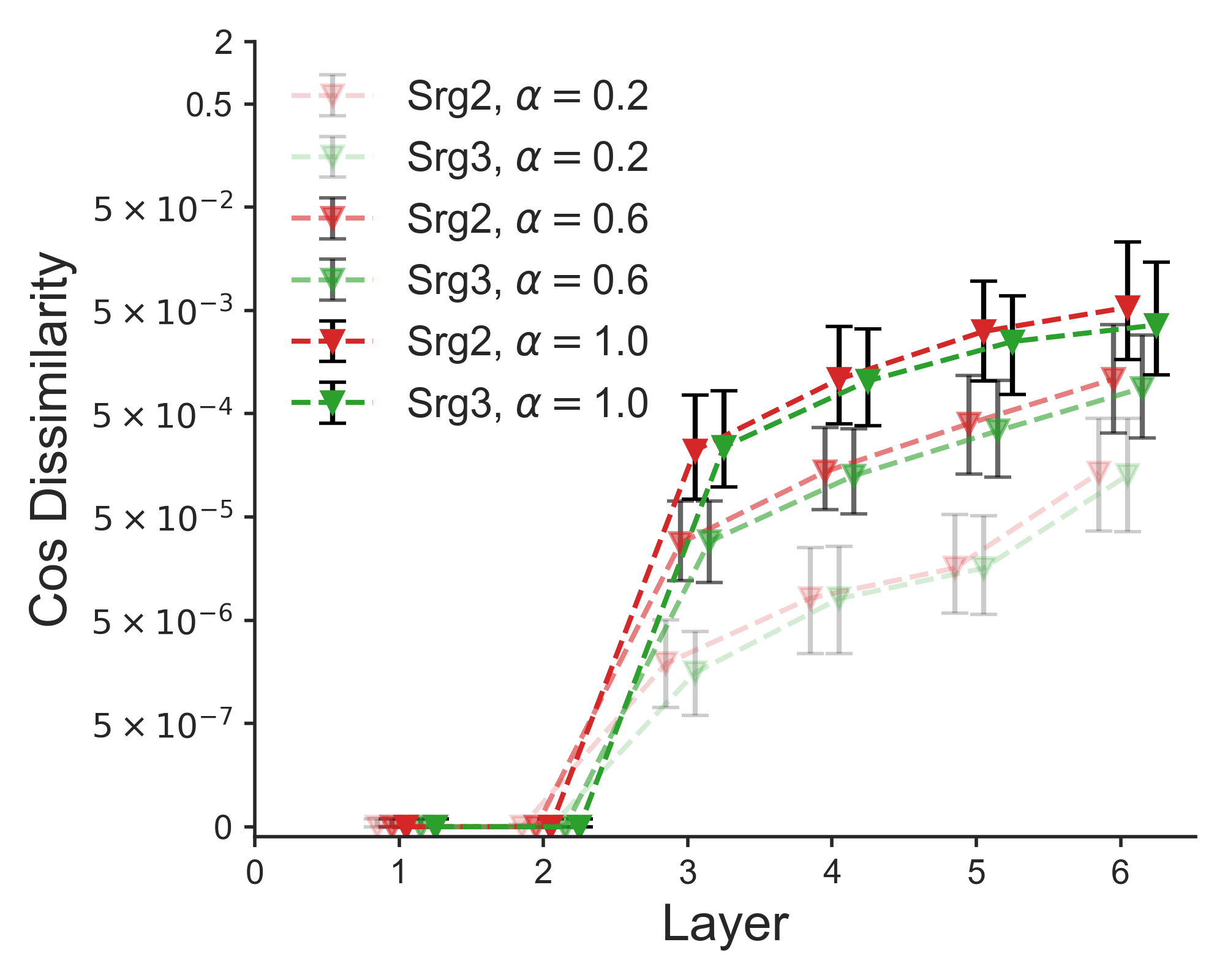}
            \includegraphics[width=0.32\linewidth]{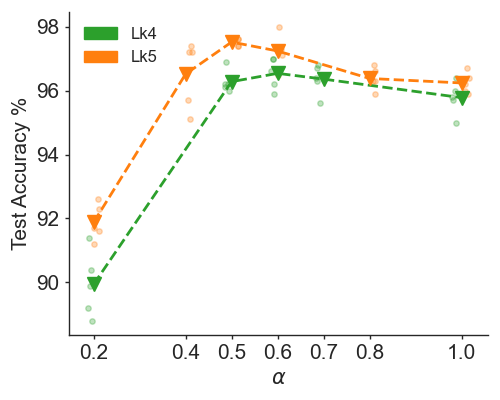}
            \includegraphics[width=0.32\linewidth]{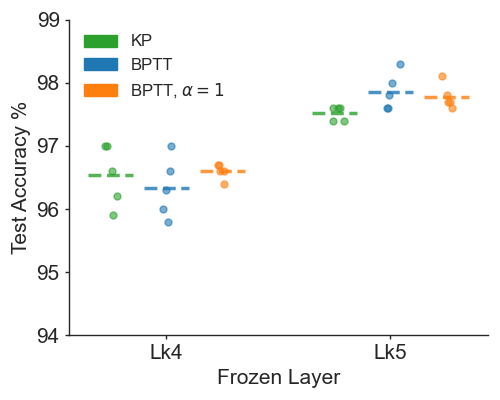}
            \caption{Further investigation of how the surrogate and nonlinearity affect K-prop approximation. (Left) Gradient cosine similarity check with BPTT like \cref{fig:GradientQuality} on $\alpha$ and another surrogate. The dissimilarity increased when bigger $\alpha$ enhance nonlinearity. Here $\hat{\rho'} = \bar{\rho'}^{\tau_\rho}$ while Srg2 uses the neurons' local $\tau$ for $\tau_\rho$ and Srg3 uses $\tau_\rho$ as a summation of downstream layers' $\tau$. Srg3 shows an advantage in early layers while slightly degrading layers close to the output. (Mid) Investigating the practical tradeoff between nonlinearity and K-prop gradient quality by tuning $\alpha$ in MNIST1D, which shows the different sweet spots on Lk4 and Lk5. (Right) Comparing with networks with $\alpha=1$ and trained with BPTT. Enhanced nonlinearity does not boost final performance.}
            \label{fig:alpha}
    \end{figure}

    \subsection{Surrogate choice}
        \label{sec:surrogate}
        The choice of surrogate derivative has an important effect on K-prop performance. In all experiments, we use the low-pass-filtered surrogate $\hat{\rho'} = \bar{\rho'}^{\tau_\rho}$ which consistently improves performance over the instantaneous choice $\hat{\rho'}=\rho'$.

        For all tasks except GSC, we set the filtering time constant $\tau_\rho$ to the local time constant $\tau$ of each neuron. For GSC, $\tau_\rho$ at each layer is instead set to the sum of the time constants in its downstream layers. This produces stronger smoothing in earlier layers, reflecting the cumulative temporal filtering along their longer paths to the output. The gradient-alignment analysis in \cref{fig:alpha} shows that this cumulative surrogate improves gradient quality in the earlier layers.
    

\newpage
\end{document}